\pdfoutput=1
\documentclass{article}

\usepackage[utf8]{inputenc} 
\usepackage[T1]{fontenc}    

\usepackage{amsmath} 
\usepackage{amsfonts} 
\usepackage[scale=1.0]{XCharter}

\usepackage[
  paper  = letterpaper,
  top    = 1.0in,
  bottom = 1.0in,
  ]{geometry}

\usepackage[usenames,dvipsnames,table]{xcolor}
\definecolor{shadecolor}{gray}{0.9}

\usepackage[final,expansion=alltext]{microtype}
\usepackage[english]{babel}
\usepackage[parfill]{parskip}
\usepackage{afterpage}
\usepackage{framed}

{\endMakeFramed}

\usepackage{lineno}

\usepackage{ragged2e}

\newcounter{parcount}

\usepackage{graphicx}
\usepackage{wrapfig}
\usepackage[labelfont=bf,font=footnotesize,width=.9\textwidth]{caption}
\usepackage[format=hang]{subcaption}

\usepackage{booktabs,multirow,multicol}       

\usepackage[algoruled,algo2e]{algorithm2e}
\usepackage{listings}
\usepackage{fancyvrb}
\fvset{fontsize=\normalsize}

\usepackage[colorlinks,linktoc=all]{hyperref}
\usepackage[all]{hypcap}
\hypersetup{citecolor=MidnightBlue}
\hypersetup{linkcolor=MidnightBlue}
\hypersetup{urlcolor=MidnightBlue}

\usepackage[nameinlink, capitalize]{cleveref}

\usepackage[acronym,smallcaps,nowarn]{glossaries}[=v4.46]

\lstdefinestyle{mystyle}{
    commentstyle=\color{OliveGreen},
    keywordstyle=\color{BurntOrange},
    numberstyle=\tiny\color{black!60},
    stringstyle=\color{MidnightBlue},
    basicstyle=\ttfamily,
    breakatwhitespace=false,
    breaklines=true,
    captionpos=b,
    keepspaces=true,
    numbers=left,
    numbersep=5pt,
    showspaces=false,
    showstringspaces=false,
    showtabs=false,
    tabsize=2
}
\usepackage{amsthm}

\usepackage{centernot}
\usepackage{nicefrac}       
\usepackage{mathtools}
\usepackage{amsbsy}
\usepackage{amstext}
\usepackage{thmtools}
\usepackage{thm-restate}

\begingroup
    \makeatletter
    \@for\theoremstyle:=definition,remark,plain\do{%
        \expandafter\g@addto@macro\csname th@\theoremstyle\endcsname{%
            \addtolength\thm@preskip\parskip
            }%
        }
\endgroup

\crefname{lemma}{lemma}{lemmas}
\Crefname{lemma}{Lemma}{Lemmas}
\crefname{thm}{theorem}{theorems}
\Crefname{thm}{Theorem}{Theorems}
\crefname{prop}{proposition}{propositions}
\Crefname{prop}{Proposition}{Propositions}
\crefname{assumption}{assumption}{assumptions}
\crefname{assumption}{Assumption}{Assumptions}

\usepackage{booktabs}
\makeatletter
\def\adl@drawiv#1#2#3{%
        \hskip.5\tabcolsep
        \xleaders#3{#2.5\@tempdimb #1{1}#2.5\@tempdimb}%
                #2\z@ plus1fil minus1fil\relax
        \hskip.5\tabcolsep}
\newcommand{\cdashlinelr}[1]{%
  \noalign{\vskip\aboverulesep
           \global\let\@dashdrawstore\adl@draw
           \global\let\adl@draw\adl@drawiv}
  \cdashline{#1}
  \noalign{\global\let\adl@draw\@dashdrawstore
           \vskip\belowrulesep}}
\makeatother

\renewcommand{\epsilon}{\varepsilon}

\declaretheorem[style=plain,name=Theorem]{theorem}

\declaretheorem[style=plain,sibling=theorem,name=Proposition]{proposition}

\declaretheorem[style=definition,sibling=theorem,name=Example]{example}

\newenvironment{example*}
 {\pushQED{\qed}\example}
 {\popQED\endexample}
\numberwithin{equation}{section}

\DeclarePairedDelimiterX\Set[1]{\lbrace}{\rbrace}%
{  #1 }

\usepackage[utf8]{inputenc} 
\usepackage[T1]{fontenc}    
\usepackage{hyperref}       
\usepackage{url}            
\usepackage{booktabs}       
\usepackage{amsfonts}       
\usepackage{nicefrac}       
\usepackage{microtype}      
\usepackage[table]{xcolor}  
\usepackage{amsmath}
\usepackage{amsthm}
\usepackage{wrapfig,lipsum,booktabs}
\usepackage{multirow}
\usepackage{subcaption}
\usepackage{algorithm}
\usepackage{algpseudocode}

\definecolor{mydarkblue}{rgb}{0,0.08,0.45}

\hypersetup{ %
    pdftitle={},
    pdfauthor={},
    pdfsubject={},
    pdfkeywords={},
    pdfborder=0 0 0,
    pdfpagemode=UseNone,
    colorlinks=true,
    linkcolor=mydarkblue,
    citecolor=mydarkblue,
    filecolor=mydarkblue,
    urlcolor=mydarkblue,
    pdfview=FitH
}

\usepackage{tcolorbox}
\tcbuselibrary{skins}
\usepackage{afterpage}
\definecolor{linen}{RGB}{250,240,230}
\usepackage{fontawesome5}
\definecolor{linen}{RGB}{250,240,230}
\usepackage{enumitem}
\usepackage{fontawesome5}
\usepackage{caption}

\crefname{appsec}{appendix}{appendices}
\Crefname{appsec}{Appendix}{Appendices}

\usepackage{appendix}
\usepackage{array}
\newcolumntype{x}[1]{>{\centering\arraybackslash}p{#1pt}}
\newcolumntype{y}[1]{>{\raggedright\arraybackslash}p{#1pt}}

\usepackage{csquotes}
\usepackage[%
minnames=1,maxnames=99,maxcitenames=2,
style=alphabetic,
doi=false,
url=false,
giveninits=true,
hyperref,
natbib,
backend=bibtex,
sorting=nyt,
backref=true
]{biblatex}%
\renewbibmacro{in:}{%
  \ifentrytype{article}{}{\printtext{\bibstring{in}\intitlepunct}}}

\renewbibmacro*{journal}{%
  \iffieldundef{journaltitle}
    {}
    {\printtext[journaltitle]{%
       \printfield[noformat]{journaltitle}%
       \setunit{\subtitlepunct}%
       \printfield[noformat]{journalsubtitle}}}}

\DeclareFieldFormat{sentencecase}{\MakeSentenceCase*{#1}}

\renewbibmacro*{title}{%
  \ifthenelse{\iffieldundef{title}\AND\iffieldundef{subtitle}}
    {}
    {\ifthenelse{\ifentrytype{article}\OR\ifentrytype{inbook}%
      \OR\ifentrytype{incollection}\OR\ifentrytype{inproceedings}%
      \OR\ifentrytype{inreference}\OR\ifentrytype{misc}}
      {\printtext[title]{%
        \printfield[sentencecase]{title}%
        \setunit{\subtitlepunct}%
        \printfield[sentencecase]{subtitle}}}%
      {\printtext[title]{%
        \printfield[titlecase]{title}%
        \setunit{\subtitlepunct}%
        \printfield[titlecase]{subtitle}}}%
     \newunit}%
  \printfield{titleaddon}}

\AtEveryBibitem{%
\ifentrytype{article}{
    \clearfield{urldate}%
    \clearfield{eprint}
    \clearfield{eid}
}{}
\ifentrytype{book}{
    \clearfield{url}%
    \clearfield{urldate}%
    \clearfield{eprint}
}{}
\ifentrytype{collection}{
    \clearfield{url}%
    \clearfield{urldate}%
    \clearfield{eprint}
}{}
\ifentrytype{incollection}{
    \clearfield{url}%
    \clearfield{urldate}%
    \clearfield{eprint}
}{}
}

\AtEveryBibitem{
    \clearfield{pages}
    \clearfield{review}%
    \clearfield{series}
    \clearfield{volume}
    \clearfield{month}
    \clearfield{isbn}
    \clearfield{issn}
    \clearlist{location}
    \clearfield{series}
    \clearlist{publisher}
    \clearname{editor}
}{}

\usepackage{thm-restate}

\crefformat{equation}{(#2#1#3)}
\crefformat{figure}{Figure~#2#1#3}
\crefname{example}{Example}{Examples}
\crefname{lemma}{Lemma}{Lemmas}
\crefname{cor}{Corollary}{Corollaries}
\crefname{theorem}{Theorem}{Theorems}
\crefname{assumption}{Assumption}{Assumptions}

\usepackage{enumitem} 
\usepackage[separate-uncertainty=true,multi-part-units=single]{siunitx} 

\declaretheoremstyle[
spacebelow=\parsep,
    spaceabove=\parsep,
  mdframed={
    backgroundcolor=gray!10!white,     
    hidealllines=true, 
    innertopmargin=8pt, 
    innerbottommargin=4pt, 
    skipabove=8pt,
    skipbelow=10pt,
    nobreak=true
}
]{grayboxed}

\crefname{gassumption}{Assumption}{Assumptions}

\usepackage{xcolor}

\definecolor{WowColor}{rgb}{.75,0,.75}
\definecolor{SubtleColor}{rgb}{0,0,.50}

\newcounter{margincounter}

\usepackage[affil-it]{authblk}

\usepackage{makecell}

\usepackage{multirow}
\usepackage{tikz}
\usetikzlibrary{positioning}
\usetikzlibrary{arrows, automata}
\usetikzlibrary{patterns}

\def\showauthornotes{1}
\ifnum\showauthornotes=1
\newcommand{\Authornote}[2]{{\sf\small\color{blue}{[#1: #2]}}}
\else
\newcommand{\Authornote}[2]{}
\fi

\title{A Latent Distribution Perspective on Evaluating and Improving Latent Generative Models}

\author{
  {\bfseries\upshape
  Xianghong Fang$^{1,*}$
  \quad
  Wenjie Shu$^{1,*}$
  \quad
  Tongda Xu$^{2}$
  } \\ \vspace{-5pt}
  {\bfseries\upshape
  Wenlong Mou$^{1}$
  \quad
  Dehan Kong$^{1}$
  \quad
  Tim G. J. Rudner$^{1,3}$
  } \\
  {\upshape
  \vspace*{10pt}
  $^1$University of Toronto \quad
  $^2$Independent\quad
  $^3$Vijil
  } \\
  {\upshape
  \vspace*{15pt}
    \begin{center}
    \href{https://sunset-clouds.github.io/Generation-Aware-Reconstruction/}{\faGlobe\enspace Website}
    \quad
    \href{https://github.com/sunset-clouds/Generation-Aware-Reconstruction}{\faGithub\enspace Code \& Models}
    \end{center}
  }
}
\date{}

\begin{document}

\maketitle

\begingroup
\renewcommand\thefootnote{}
\footnotetext{
$^{*}$ Equal contribution.
}
\endgroup

\begin{abstract}


In latent generative models, reconstruction quality is often assumed to correlate with generative performance. However, reconstruction FID (rFID) can exhibit weak or even negative correlation with generation FID (gFID). We attribute this misalignment to a latent distribution mismatch: reconstruction evaluates the decoder on encoder-induced latents, whereas generation uses the same decoder on latents produced by the generative model. To characterize this shift, we introduce generation-aware reconstruction (GAR), which constructs a continuous trajectory from standard reconstruction toward generation by perturbing encoder latents with noise and denoising them through the generative model before decoding. GAR probes the decoder behavior along this trajectory, making the transition from encoder to generation-time latent distributions observable and diagnosable. The resulting trajectory-based diagnostic, GAR-FID, exhibits strong empirical correlation with gFID across diverse tokenizers and scales. Importantly, intermediate GAR latents become more generation-aware while preserving correspondence with their source images, thereby retaining paired supervision that is absent for fully generated latents. This correspondence enables decoder adaptation on intermediate GAR latents, consistently improving generative quality across model scales. Overall, latent distribution mismatch provides a useful perspective for evaluating and improving latent generative models.
\end{abstract}


\section{Introduction}
Latent generative modeling has become a dominant paradigm for high-quality image synthesis, typically using a visual tokenizer to map images into a latent space where generation is modeled by diffusion or flow matching~\citep{Rombach2022HighResolutionIS,Preechakul2021DiffusionAT,Peebles2022ScalableDM,Albergo2023BuildingNF,Liu2023FlowSA,Geng2025MeanFF}.
Reconstruction quality is commonly used to evaluate the tokenizer and is often implicitly assumed to correlate with downstream generative performance: a decoder that better reconstructs images from encoder latents might also be expected to produce better samples during generation~\citep{Oord2017NeuralDR,Esser2020TamingTF,Rombach2022HighResolutionIS,Preechakul2021DiffusionAT}.
However, recent empirical studies~\citep{Yao2025ReconstructionVG,Kouzelis2025EQVAEER,Ye2025DistributionMV,Skorokhodov2025ImprovingTD,Chen2025MaskedAA,Xu2026MakingRF} have challenged this assumption.
In practice, reconstruction FID (rFID) can exhibit weak or even negative correlation with generation FID (gFID), suggesting a possible misalignment between reconstruction quality and generative performance.

\begin{figure*}[t]
    \centering
    \vspace*{-3pt}
    \includegraphics[width=\linewidth]{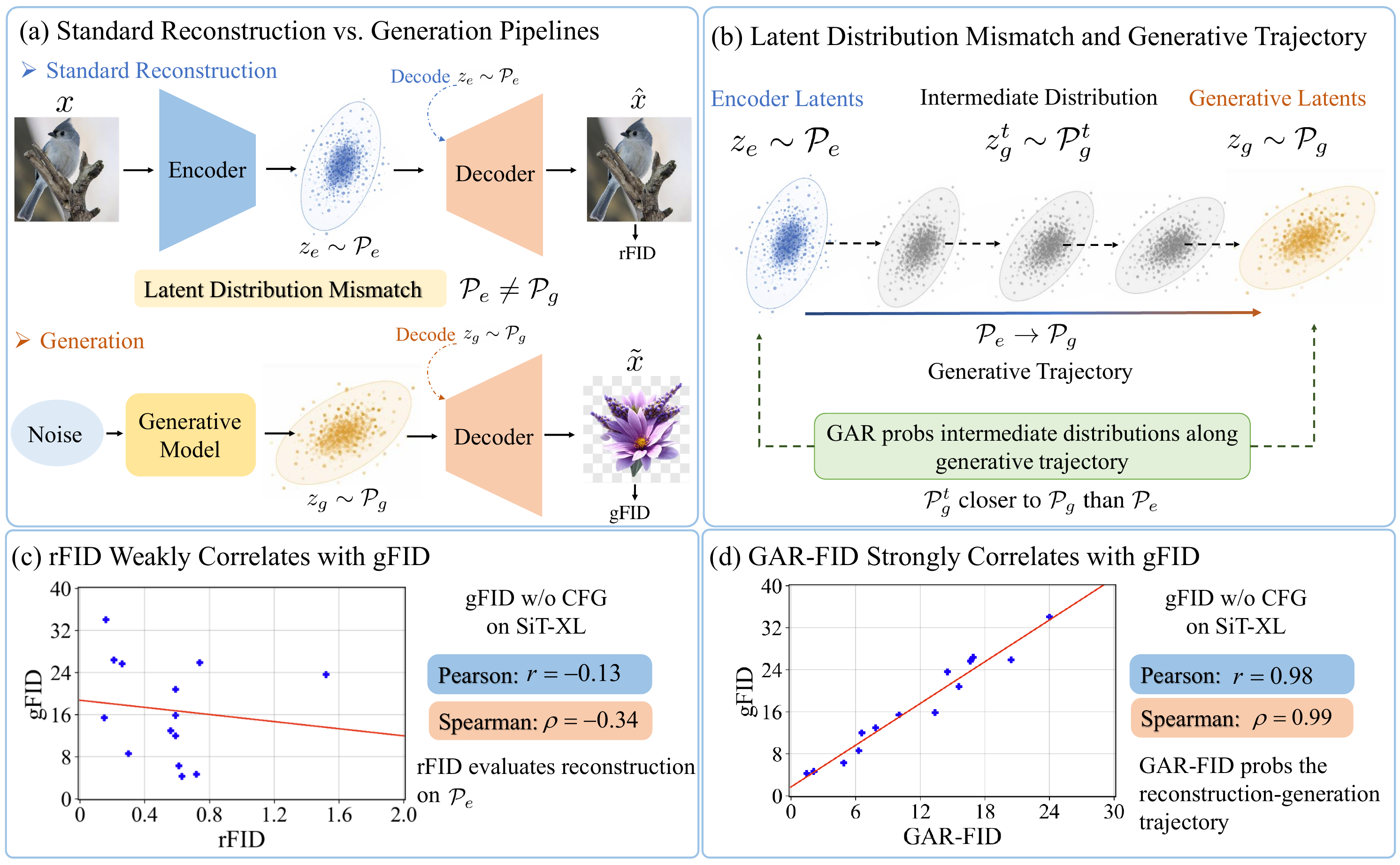}
    \vspace{-4ex}
    \caption{\small{\textbf{Generation-aware reconstruction exposes the reconstruction-generation latent transition.}
    (a) Reconstruction and generation use the same decoder on different latent distributions:
    reconstruction decodes encoder latents $z_e \sim \mathcal{P}_e$, whereas generation decodes generator-produced latents $z_g \sim \mathcal{P}_g$.
    (b) GAR uses the generative process to construct a trajectory of intermediate latent distributions $\mathcal{P}_g^t$ from reconstruction toward generation.
    (c) In this setting, rFID is weakly correlated with gFID, with Spearman rank correlation $\rho=-0.34$.
    (d) GAR-FID at $\eta_t=0.8$ exhibits strong correlation with gFID, achieving $\rho=0.99$.}}
    \label{fig:motivation}
    \vspace{-2ex}
\end{figure*}

In this paper, we investigate the causes of misalignment between reconstruction quality and generation performance in latent generative models. We hypothesize that this misalignment can be explained by structural differences between the latent distributions encountered during reconstruction and generation.
That is, the fact that reconstruction evaluates the decoder on encoder latents $z_e \sim \mathcal{P}_e$, whereas generation uses the same decoder on generative latents $z_g \sim \mathcal{P}_g$, as illustrated in Figure~\ref{fig:motivation}(a).
When $\mathcal{P}_e$ and $\mathcal{P}_g$ differ, decoder behavior under encoder latents need not reflect its behavior at generation time, potentially explaining why reconstruction quality may not closely track generative performance.
Moreover, comparing only the two endpoints leaves unobserved how decoder behavior changes as its inputs shift from encoder to generation-time latent distributions.

To expose this transition, we introduce \emph{generation-aware reconstruction} (GAR), which constructs a continuous trajectory from standard reconstruction toward generation, as illustrated in Figure~\ref{fig:motivation}(b).
Specifically, GAR perturbs encoder latents with noise and denoises them through the generative model before decoding, producing intermediate latent distributions between the reconstruction and generation endpoints.
By probing the decoder along this trajectory, GAR makes the reconstruction-generation transition traceable.
The resulting trajectory-based diagnostic, GAR-FID, exhibits high correlation with gFID across diverse tokenizers, generator architectures, and model scales---in contrast to the weak association observed for rFID (Figure~\ref{fig:motivation}(c,d)).
As such, GAR complements endpoint measurements by revealing how decoder behavior evolves between reconstruction and generation.

Finally, the GAR trajectory also provides a natural opportunity for improving generation.
At intermediate noise levels, GAR latents move toward generation-time distributions while retaining correspondence with their source images, thereby providing generation-aware paired supervision.
Based on this insight, we propose a simple \emph{decoder adaptation} approach that freezes the encoder and generative model and fine-tunes only the decoder on intermediate GAR latents using standard tokenizer objectives.
This generation-oriented adaptation exposes the decoder to latent distributions closer to those encountered during generation while preserving meaningful image-level supervision, leading to consistent improvements in generative quality across model scales.
Overall, our results establish latent distribution mismatch as a useful perspective for evaluating and improving latent generative models.

Our contributions are summarized as follows:
\vspace*{-3pt}
\begin{enumerate}[topsep=0pt, align=left, leftmargin=15pt, labelindent=1pt,
listparindent=\parindent, labelwidth=0pt, itemindent=!, itemsep=3pt, parsep=0pt]
\item We identify a \textbf{latent distribution mismatch} between reconstruction and generation: the same decoder is evaluated on encoder latents during reconstruction but operates on generator-produced latents during generation.
This provides a distributional explanation for why reconstruction quality need not track generative performance.

\item We introduce \textbf{generation-aware reconstruction (GAR)}, which constructs and probes the reconstru\\ction-generation latent trajectory.
GAR-FID quantifies decoder behavior along this trajectory and provides a reliable evaluation proxy for generative performance, exhibiting strong correlation with gFID across diverse tokenizer and generator configurations.

\item We show that this trajectory provides a principled route from \textbf{diagnosis to adaptation}: intermediate GAR latents move toward generation-time distributions while retaining source correspondence, providing paired, generation-aware supervision for decoder adaptation and consistently improving generative quality across model scales.
\end{enumerate}
\section{Revisiting Reconstruction and Generation}
\label{sec:reconstruction and generation pipeline}
Latent generative models couple a tokenizer with a generative model through a shared decoder, yet the two are trained independently and operate on different latent distributions. This section formalizes this distributional gap and introduces generation-aware reconstruction (GAR) as a framework for probing the transition between reconstruction and generation.

\subsection{Latent Distribution Mismatch Between Reconstruction and Generation}
\label{sec:standard_reconstruction}

Latent visual generative models~\citep{Rombach2022HighResolutionIS} are typically composed of two separately trained components: a continuous visual tokenizer~\citep{Chen2024SoftVQVAEE1,Chen2025MaskedAA,yang2026latent,Kouzelis2025EQVAEER} and a latent diffusion or flow-matching model~\citep{Preechakul2021DiffusionAT,Peebles2022ScalableDM,Albergo2023BuildingNF,Liu2023FlowSA,Geng2025MeanFF}.

As illustrated in Figure~\ref{fig:motivation}(a), standard reconstruction encodes an input image $x$ to a latent $z_e=\mathcal{E}_{\theta}(x)\sim\mathcal{P}_e$ and decodes it as $\hat{x}=\mathcal{D}_{\phi}(z_e)$; quality is measured by Fr'echet Inception Distance~\citep[FID;][]{Heusel2017GANsTB}, referred to as rFID. Generation starts from noise $\epsilon\sim\mathcal{N}(0,I)$: a generative model $\mathcal{G}_{\varphi}$ produces a latent $z_g=\mathcal{G}_{\varphi}(\epsilon)\sim\mathcal{P}_g$, decoded as $\tilde{x}=\mathcal{D}_{\phi}(z_g)$, with quality measured by gFID.

The two pathways differ in the distribution of latents the decoder receives: $\mathcal{P}_e$ in reconstruction, $\mathcal{P}_g$ in generation. Good decoder behavior under one does not guarantee good behavior under the other, which is why rFID and gFID can diverge. Both metrics capture only their respective endpoints; how decoder behavior evolves between them remains uncharacterized.

\subsection{Generation-Aware Reconstruction}
\label{sec:GAR}

\begin{wrapfigure}[12]{r}{0.55\textwidth}
\small
\vspace{-4.5mm}
\begin{center}
\includegraphics[width=0.55\textwidth]{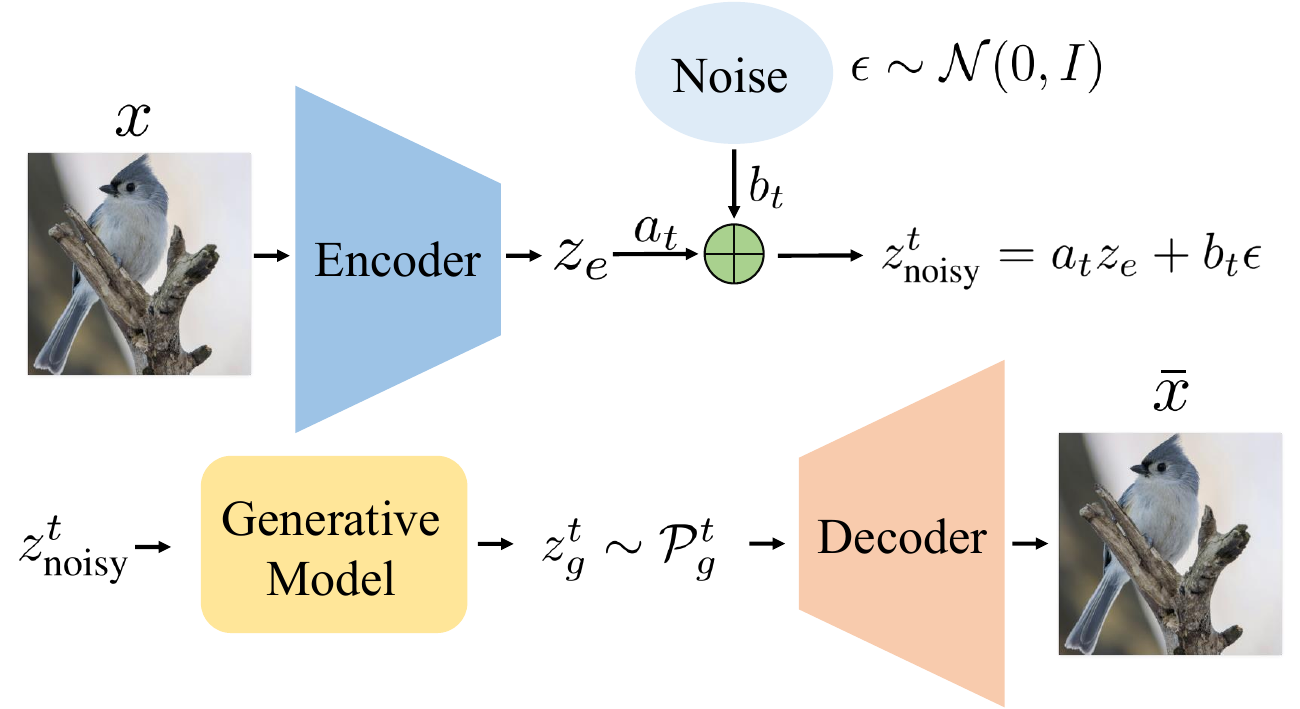}
\vspace{-7mm}
\caption{\small{Illustration of the GAR process.}}
\label{fig:gar pipeline}
\end{center}
\end{wrapfigure}
To characterize this transition, we introduce \emph{generation-aware reconstruction} (GAR), which constructs a controlled trajectory from standard reconstruction toward generation. As illustrated in Figure~\ref{fig:gar pipeline}, given an encoder latent $z_e\sim\mathcal{P}_e$, GAR perturbs it with noise, yielding $z^t_{\mathrm{noisy}}=a_t z_e+b_t\epsilon$, where $\epsilon\sim\mathcal{N}(0,I)$ and $a_t,b_t$ are the noise-schedule coefficients at timestep $t$. We define the relative noise level as $\eta_t=b_t/(a_t+b_t)$. The generative model then denoises $z^t_{\mathrm{noisy}}$ from timestep $t$, producing $z_g^t=\mathcal{G}_{\varphi}(z^t_{\mathrm{noisy}},t)\sim\mathcal{P}_g^t$, decoded as $\bar{x}=\mathcal{D}_{\phi}(z_g^t)$. The noise level $\eta_t$ determines the position along the reconstruction-generation trajectory. At $\eta_t=0$, no perturbation or denoising is applied, so $z_g^t=z_e$ and $\mathcal{P}_g^t=\mathcal{P}_e$, recovering standard reconstruction. At $\eta_t=1$, the encoder latent is fully replaced by noise and the full generative process is applied, so $z_g^t=z_g$ and $\mathcal{P}_g^t=\mathcal{P}_g$, recovering the generation endpoint. As $\eta_t$ increases, the intermediate distributions $\mathcal{P}_g^t$ progressively move from $\mathcal{P}_e$ toward $\mathcal{P}_g$, while source correspondence gradually weakens. GAR thus connects the two endpoints through a controlled latent-distribution trajectory.

We define GAR-FID as the FID between source images and their GAR outputs $\bar{x}$ at a specified noise level $\eta_t$. By varying $\eta_t$, GAR-FID tracks how decoder behavior changes as its inputs move along the generator-induced trajectory from $\mathcal{P}_e$ toward $\mathcal{P}_g$: rFID characterizes the reconstruction endpoint, GAR-FID probes the intermediate distributions $\mathcal{P}_g^t$, and gFID characterizes the generation endpoint. As shown in Section~\ref{sec:rfid_gfid}, GAR-FID becomes strongly associated with gFID as the trajectory approaches the generation-time regime across diverse tokenizer and generator configurations.

At intermediate noise levels, GAR latents can become more generation-aware while retaining source correspondence. This paired structure allows the same trajectory used for diagnosis to also provide a training distribution for decoder adaptation, as described in Section~\ref{sec:method}.

\section{Characterizing the Reconstruction-Generation Latent Shift}
\label{sec:analyses on IMF}
We now empirically examine the latent distribution perspective developed in Section~\ref{sec:reconstruction and generation pipeline}. We first quantify the mismatch between $\mathcal{P}_e$ and $\mathcal{P}_g$. We then verify that GAR induces the expected shift from $\mathcal{P}_e$ toward $\mathcal{P}_g$ as the noise level increases. Finally, we study how the endpoint mismatch is associated with generative performance. Additional analyses of structural and perceptual reconstruction quality along the GAR trajectory are provided in Appendix~\ref{appendix:structural-perceptual}.

\subsection{Latent Distribution Mismatch}
\label{sec:latent mismatch}

\begin{wrapfigure}[15]{r}{0.60\textwidth}
\small
\vspace{-5.5mm}
\begin{center}
\includegraphics[width=0.60\textwidth]{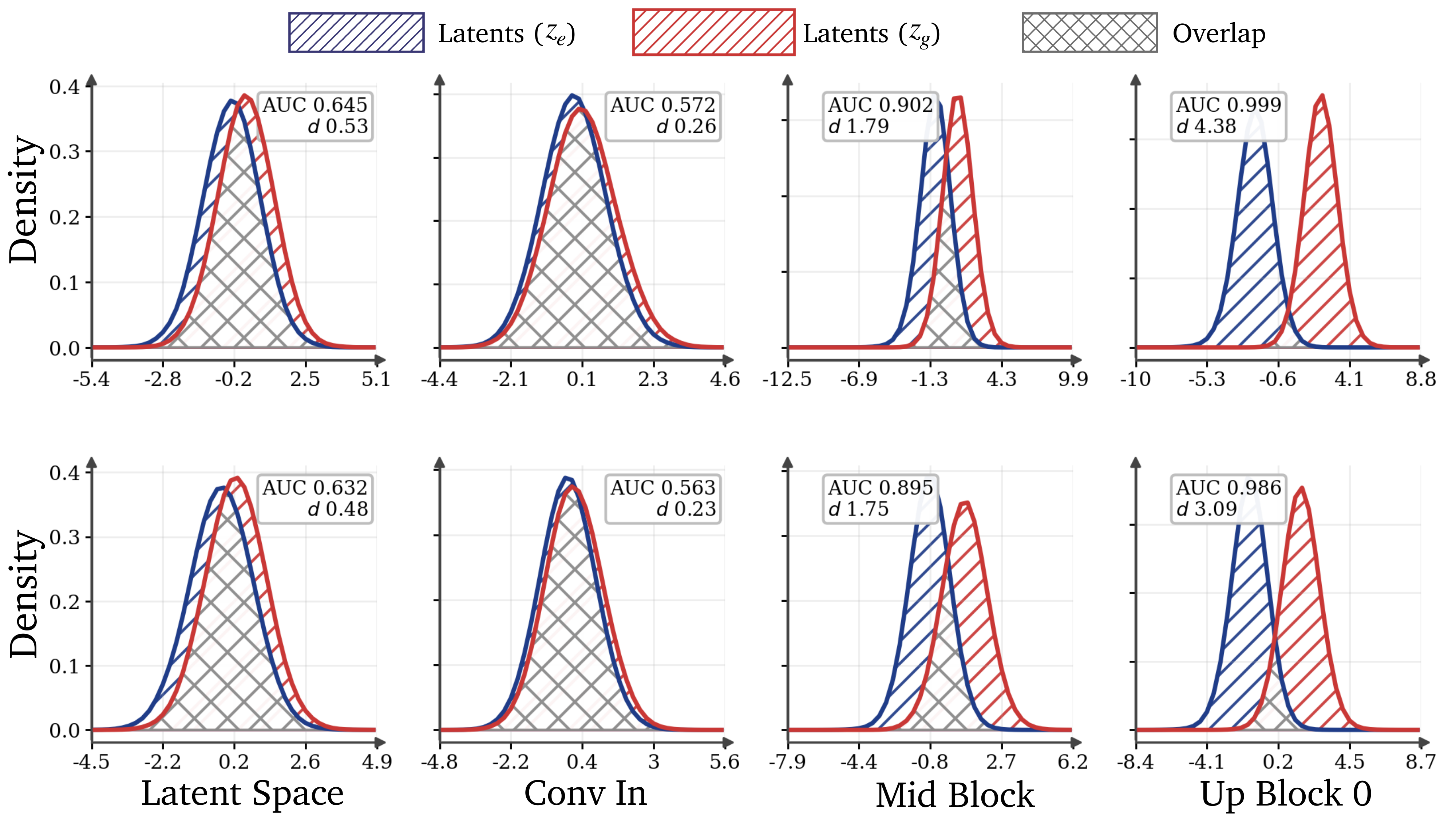}
\vspace{-6mm}
\caption{\small{LDA projections of $z_e$ and $z_g$ for iMF-B/2 (top) and iMF-XL/2 (bottom). The vertical axis shows the Gaussian-smoothed, normalized histogram density of the one-dimensional LDA scores.
}}
\label{fig:lda}
\end{center}
\vspace{-2mm}
\end{wrapfigure}

To characterize the distribution mismatch between encoder latents $z_e$ and generative latents $z_g$, we conduct a systematic empirical analysis using the SD-VAE~\citep{Rombach2022HighResolutionIS} encoder and iMF~\citep{Geng2025ImprovedMF} as representative models. Since each latent is high-dimensional ($z_e\in \mathbb{R}^{32 \times 32 \times 4}$, i.e., 4096 dimensions), conventional low-dimensional visualization techniques such as 2D PCA were not sufficiently informative in our analyses. To overcome this limitation, we adopt Linear Discriminant Analysis (LDA) to identify the direction that maximizes separation between the two distributions, and visualize the resulting one-dimensional projections as Gaussian-smoothed densities, as shown in Figure~\ref{fig:lda}.

Our results reveal that the distribution gap between generative and encoder latents is smaller for iMF-XL/2 compared to iMF-B/2. This trend is further corroborated by a high-dimensional latent Fréchet distance (Latent-FD, see Appendix~\ref{appendix:latent-fd}), which yields a lower distance for iMF-XL/2 (91.55) than for iMF-B/2 (103.64), consistent with lower LDA AUC and reduced differences in distributional means. These results show that, within this controlled iMF comparison, the larger generative model exhibits a smaller measured mismatch. Furthermore, separability increases systematically in progressively deeper decoder feature spaces. At Decoder Up Block 0, the LDA AUC reaches 0.999, indicating near-complete separation between encoder- and generator-induced representations. We refer to this phenomenon as the \emph{decoder amplification effect}: differences that are moderate in latent space become substantially more pronounced in downstream decoder representations. This observation suggests that the decoder can amplify latent distribution mismatch as representations propagate toward image space.

\subsection{Distribution Shift Along the Reconstruction-Generation Trajectory}
\label{sec:progressive distribution shift}

\begin{wrapfigure}[11]{r}{0.41\textwidth}
\small
\vspace{-4.5mm}
\begin{center}
\includegraphics[width=0.41\textwidth]{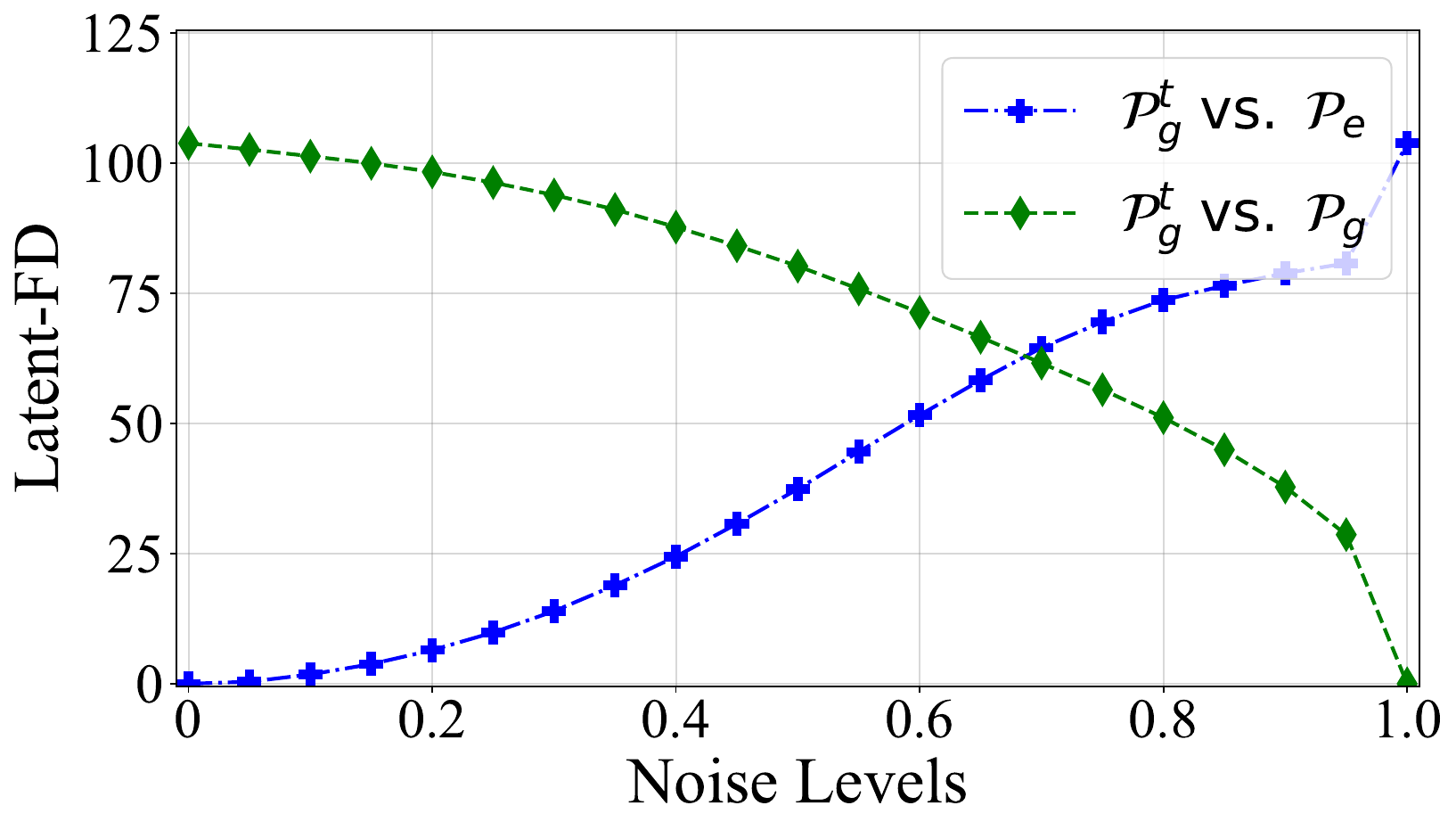}
\vspace{-7mm}
\caption{\small{Latent-FD from intermediate GAR distributions $\mathcal{P}_g^t$ to the reconstruction endpoint $\mathcal{P}_e$ and generation endpoint $\mathcal{P}_g$. 
}}
\label{fig:latent-fd}
\end{center}
\vspace{-2mm}
\end{wrapfigure} 
We next verify the latent distribution shift induced by GAR by measuring the discrepancy between $\mathcal{P}_g^t$ and the two endpoint distributions $\mathcal{P}_e$ and $\mathcal{P}_g$. Using the SD-VAE encoder~\citep{Rombach2022HighResolutionIS} and iMF-B/2~\citep{Geng2025ImprovedMF}, we extract GAR latents on ImageNet~\citep{Deng2009ImageNetAL} and compute the corresponding Latent-FD.

As shown in Figure~\ref{fig:latent-fd}, increasing the noise level produces a consistent distributional shift: the Latent-FD between $\mathcal{P}_g^t$ and $\mathcal{P}_e$ increases, while that between $\mathcal{P}_g^t$ and $\mathcal{P}_g$ decreases. This confirms that GAR progressively moves decoder inputs away from the reconstruction regime and toward the generation-time regime, consistent with the reconstruction-generation trajectory defined in Section~\ref{sec:GAR}. As source correspondence weakens along this trajectory, structural and perceptual reconstruction quality also gradually deteriorates; detailed results are provided in Appendix~\ref{appendix:structural-perceptual}.

\subsection{Latent Distribution Mismatch and Generative Performance}
\label{sec:distribution matching-generative performance}
We finally examine whether the endpoint mismatch between $\mathcal{P}_e$ and $\mathcal{P}_g$ is associated with generative performance. We measure the mismatch using Latent-FD and compare it with gFID, both with and without classifier-free guidance (CFG). Our evaluation uses the SD-VAE tokenizer~\citep{Rombach2022HighResolutionIS} with iMF-B/2, iMF-M/2, and iMF-XL/2~\citep{Geng2025ImprovedMF} on ImageNet~\citep{Deng2009ImageNetAL}.

\begin{wrapfigure}[9]{r}{0.70\textwidth}
\small
\vspace{-4.5mm}
\begin{center}
\includegraphics[width=0.70\textwidth]{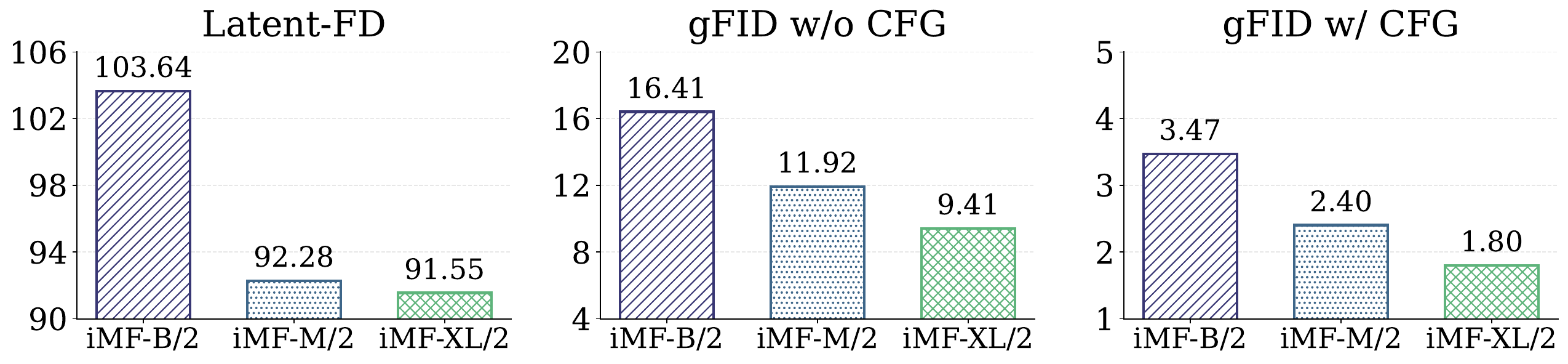}
\vspace{-6mm}
\caption{\small{Latent distribution mismatch and generative performance across iMF model scales. Smaller Latent-FD between $\mathcal{P}_e$ and $\mathcal{P}_g$ coincides with lower gFID, both without and with CFG.}}
\label{fig:bar}
\end{center}
\vspace{-2mm}
\end{wrapfigure}
As shown in Figure~\ref{fig:bar}, models with smaller Latent-FD consistently achieve lower gFID across the three model scales, both with and without CFG. This trend suggests a close association between latent distribution mismatch and generative performance. One possible explanation is that a larger mismatch places the decoder on generation-time latents farther from the distribution encountered during reconstruction, potentially degrading sample quality. Motivated by this empirical relationship, we next provide a theoretical perspective showing how latent distribution mismatch can contribute to an upper bound on generative error.

\begin{proposition}[A Lipschitz Bound Linking Latent-FD and gFID ]
\label{prop:latent-fd-gfid-bound}
Let $\psi:\mathcal{X}\to\mathbb{R}^{2048}$ denote the Inception feature extractor used for FID, and define the decoder-feature map $T = \psi\circ\mathcal{D}_{\phi}$. Suppose that $\mathcal{P}_e$ and $\mathcal{P}_g$ are Gaussian latent distributions. Assume that the decoder $\mathcal{D}_{\phi}$ and feature extractor $\psi$ are Lipschitz with constants $L_{\mathcal{D}}$ and $L_{\psi}$, respectively. Then we have
\begin{equation}
\sqrt{\mathrm{gFID}}
\leq
\sqrt{\mathrm{rFID}}
+ L_{\psi}L_{\mathcal{D}}\sqrt{\mathrm{Latent\text{-}FD}(\mathcal{P}_e,\mathcal{P}_g)}.
\vspace{-1ex}
\end{equation}
\end{proposition}
The proof is provided in Appendix~\ref{app:proof of proposition}. Proposition~\ref{prop:latent-fd-gfid-bound} provides a theoretical lens for the empirical trend in Figure~\ref{fig:bar}. Under the stated Gaussian and Lipschitz assumptions, gFID is upper-bounded by the reconstruction term together with a contribution that grows with latent distribution mismatch and is scaled by the decoder-feature Lipschitz constant. The bound does not imply that reducing Latent-FD must reduce gFID; rather, it identifies latent mismatch as one factor that can enlarge the gap between reconstruction and generation, consistent with the empirical association above.
\section{Tracking Decoder Behavior from Reconstruction to Generation}
\label{sec:rfid_gfid}

We now examine how decoder behavior evolves along the reconstruction-generation trajectory and how GAR-FID captures this evolution. We first use a controlled setting with a fixed tokenizer to isolate variation induced by the generative model and connect GAR-FID to the underlying latent distribution shift. We then examine whether the relationship between GAR-FID and generative performance remains stable when both tokenizers and generators vary.

\subsection{Controlled Study With a Fixed Tokenizer}
\label{sec:controlled_analysis}

\begin{wrapfigure}[9]{r}{0.70\textwidth}
\small
\vspace{-4.5mm}
\begin{center}
\includegraphics[width=0.70\textwidth]{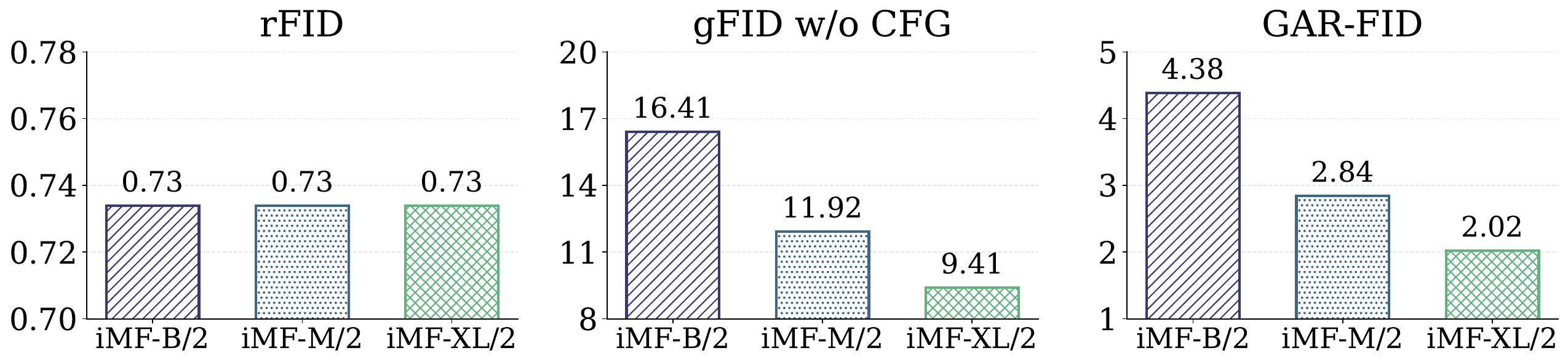}
\vspace{-6mm}
\caption{\small{Metric behavior under a fixed tokenizer (SD-VAE) as generator scale increases from iMF-B/2 to iMF-XL/2. rFID remains unchanged by construction, while gFID and GAR-FID vary with the generative model.}}
\label{fig:fid comparison}
\end{center}
\vspace{-2mm}
\end{wrapfigure}

We begin with a controlled study that fixes the SD-VAE tokenizer~\citep{Rombach2022HighResolutionIS} while varying the generative model across three scales: iMF-B/2, iMF-M/2, and iMF-XL/2~\citep{Geng2025ImprovedMF}. Fixing the tokenizer places all latent distributions in a shared latent space, making $\mathrm{Latent\text{-}FD}(\mathcal{P}_e,\mathcal{P}_g)$ directly comparable across generators and isolating variation induced by the generative model. Figure~\ref{fig:fid comparison} shows that gFID decreases substantially with increasing generator scale, while rFID remains unchanged because the tokenizer is fixed. This invariance is expected: rFID characterizes reconstruction under $\mathcal{P}_e$ and does not depend on the generative model. GAR-FID, in contrast, varies consistently with generator scale because its intermediate distributions $\mathcal{P}_g^t$ depend on the corresponding generator. This controlled setting therefore highlights their distinct roles: rFID characterizes tokenizer reconstruction, whereas GAR-FID reflects generator-induced variation along the reconstruction-generation trajectory.

\paragraph{Connecting GAR-FID to Latent Distribution Mismatch.} Section~\ref{sec:distribution matching-generative performance} showed that, within this fixed-tokenizer setting, a larger endpoint mismatch between $\mathcal{P}_e$ and $\mathcal{P}_g$ is empirically associated with worse generative performance. We summarize this observed relationship as $
\mathrm{Latent\text{-}FD}(\mathcal{P}_e,\mathcal{P}_g)\uparrow
\;\leadsto\;
\mathrm{gFID}\uparrow
$. Hereafter, $\leadsto$ denotes an empirical relationship rather than a causal implication.

\begin{wrapfigure}[10]{r}{0.42\textwidth}
\small
\vspace{-4.5mm}
\begin{center}
\includegraphics[width=0.42\textwidth]{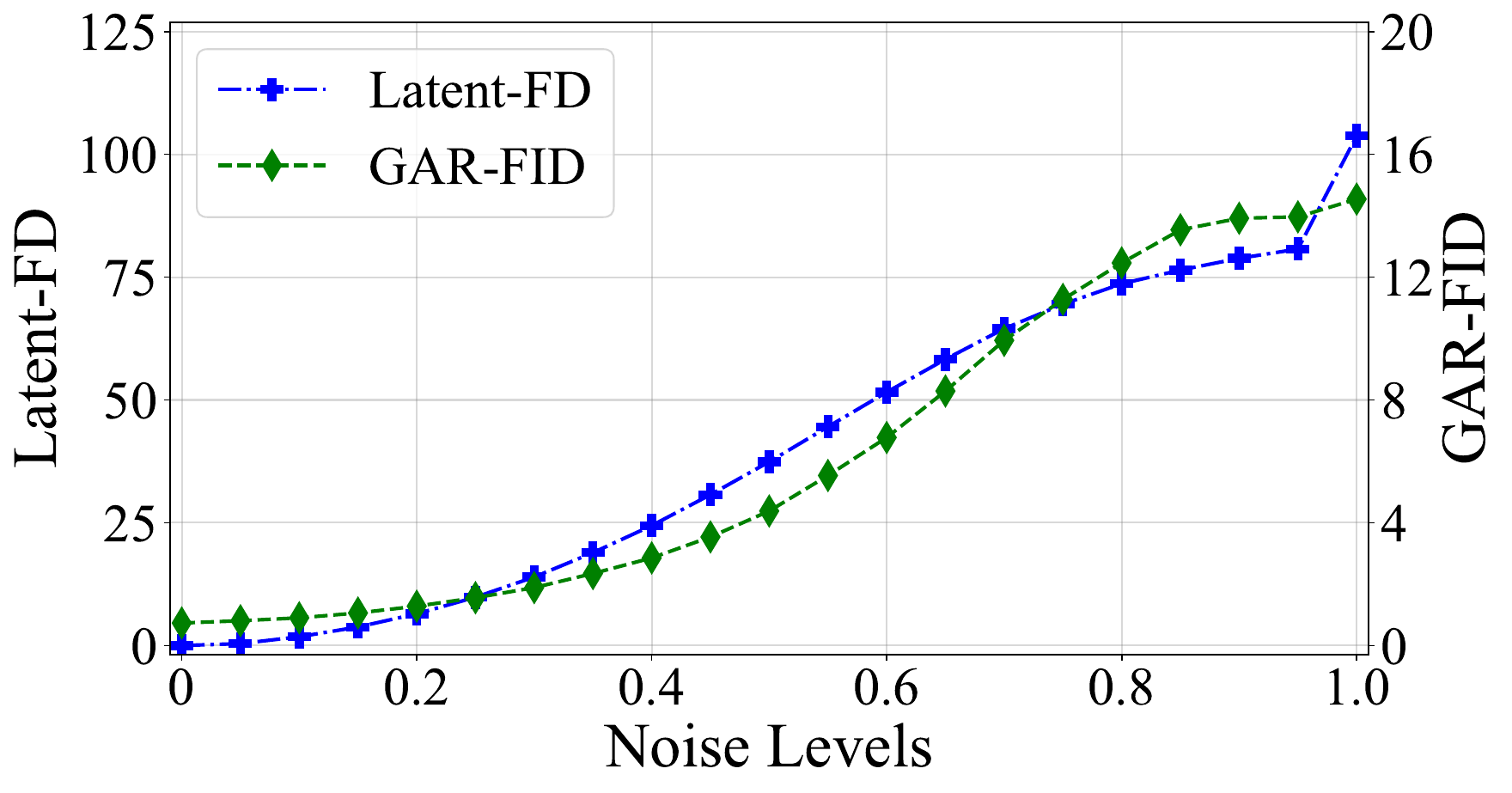}
\vspace{-6mm}
\caption{\small{GAR-FID (w/o CFG) closely tracks $\mathrm{Latent\text{-}FD}(\mathcal{P}_g^t,\mathcal{P}_e)$ across noise levels.}}
\label{fig:gar-fid-latent-fd}
\end{center}
\vspace{-2mm}
\end{wrapfigure}
We next examine how GAR-FID connects to this endpoint relationship through the intermediate distributions $\mathcal{P}_g^t$. As shown in Figure~\ref{fig:gar-fid-latent-fd}, GAR-FID exhibits near-perfect rank correlation with $\mathrm{Latent\text{-}FD}(\mathcal{P}_g^t,\mathcal{P}_e)$ across the evaluated noise levels (Spearman's $\rho=1.0$, $p<10^{-8}$). Thus, as GAR progressively shifts decoder inputs away from $\mathcal{P}_e$, GAR-FID changes systematically with the underlying latent shift: $\mathrm{GAR\text{-}FID}\uparrow
\leadsto
\mathrm{Latent\text{-}FD}(\mathcal{P}_g^t,\mathcal{P}_e)\uparrow$. Moreover, under the fixed-tokenizer setting, $\mathrm{Latent\text{-}FD}(\mathcal{P}_g^t,\mathcal{P}_e)$ preserves the model ordering of the endpoint mismatch $\mathrm{Latent\text{-}FD}(\mathcal{P}_g,\mathcal{P}_e)$ across noise levels (Appendix~\ref{appendix:intermediate_latent_fd}). We summarize this ordering relationship as
$
\mathrm{Latent\text{-}FD}(\mathcal{P}_g^t,\mathcal{P}_e)\uparrow
\;\leadsto\;
\mathrm{Latent\text{-}FD}(\mathcal{P}_g,\mathcal{P}_e)\uparrow .
$

Combining these observations yields the following empirical association chain:
\begin{equation}
\mathrm{GAR\text{-}FID}\uparrow
\;\leadsto\;
\mathrm{Latent\text{-}FD}(\mathcal{P}_g^t,\mathcal{P}_e)\uparrow
\;\leadsto\;
\mathrm{Latent\text{-}FD}(\mathcal{P}_g,\mathcal{P}_e)\uparrow
\;\leadsto\;
\mathrm{gFID}\uparrow .
\label{eq:gar-association-chain}
\end{equation}
The three links respectively capture the correlation between GAR-FID and intermediate latent mismatch, the preservation of generator ordering from intermediate to endpoint mismatch, and the empirical association between endpoint mismatch and generative performance. This chain provides an empirical explanation for why GAR-FID becomes increasingly associated with gFID as the trajectory approaches the generation-time regime.

\paragraph{Beyond the Controlled Setting.}
Because the tokenizer is fixed, all latent distributions share the same representation space and Latent-FD can be compared directly across generators. When tokenizers vary, their latent spaces differ and Latent-FD is no longer directly comparable. We therefore turn to metric correlations to assess whether the GAR-FID--gFID relationship remains stable across tokenizer and generator variations.

\subsection{Robustness Across Tokenizers and Generators}
\label{sec:generalization_analysis}

We next examine whether the relationship between GAR-FID and gFID remains stable when both the tokenizer and generative model vary. Following iFID~\citep{Xu2026MakingRF}, we evaluate 13 tokenizers with two latent generative models, SiT-B and SiT-XL~\citep{Ma2024SiTEF}, and compute rFID, GAR-FID, and gFID for each tokenizer-generator configuration. All metrics are reported in Table~\ref{tab:sit-b} and Table~\ref{tab:sit-xl}, with detailed experimental settings provided in Appendix~\ref{appendix:additional_generalization_results}.

\begin{table}[!t]
\caption{\small{Correlation between FID-based metrics and generative performance (gFID). Correlations are computed across models obtained by training SiT with 13 pre-trained VAEs. We report Pearson correlation (PCC) and Spearman rank correlation (SRCC), with and without CFG, for SiT-B, SiT-XL, and their mixed configurations. The correlation between GAR-FID and gFID strengthens as the noise level $\eta_t$ increases, while rFID remains negatively correlated with gFID across settings. \textbf{Bold} indicates the best results. $^{\dagger}$ denotes results from~\citep{Xu2026MakingRF}, and $^{\star}$ indicates our reproduced results. Full results are provided in Table~\ref{tab:resultfid-full} in Appendix~\ref{appendix:additional_generalization_results}.}}
\vspace{-2ex}
\label{tab:resultfid}
\resizebox{\linewidth}{!}{
\begin{tabular}{@{}lcccccccccccc@{}}
\toprule
\multirow{3}{*}{Metrics} & \multicolumn{4}{c}{gFID SiT/B} & \multicolumn{4}{c}{gFID SiT/XL} & \multicolumn{4}{c}{gFID SiT/mixed}\\
\cmidrule(lr){2-5}\cmidrule(lr){6-9}\cmidrule(lr){10-13}
 & \multicolumn{2}{c}{w/o CFG} & \multicolumn{2}{c}{w/ CFG} & \multicolumn{2}{c}{w/o CFG} & \multicolumn{2}{c}{w/ CFG} & \multicolumn{2}{c}{w/o CFG} & \multicolumn{2}{c}{w/ CFG} \\
\cmidrule(lr){2-3}\cmidrule(lr){4-5}\cmidrule(lr){6-7}\cmidrule(lr){8-9}\cmidrule(lr){10-11}\cmidrule(lr){12-13}
 & PCC ({\color{green!60!black}\bfseries$\pmb{\uparrow}$}) & SRCC ({\color{green!60!black}\bfseries$\pmb{\uparrow}$}) & PCC ({\color{green!60!black}\bfseries$\pmb{\uparrow}$}) & SRCC ({\color{green!60!black}\bfseries$\pmb{\uparrow}$}) & PCC ({\color{green!60!black}\bfseries$\pmb{\uparrow}$}) & SRCC ({\color{green!60!black}\bfseries$\pmb{\uparrow}$}) & PCC ({\color{green!60!black}\bfseries$\pmb{\uparrow}$}) & SRCC ({\color{green!60!black}\bfseries$\pmb{\uparrow}$}) & PCC ({\color{green!60!black}\bfseries$\pmb{\uparrow}$}) & SRCC ({\color{green!60!black}\bfseries$\pmb{\uparrow}$}) & PCC ({\color{green!60!black}\bfseries$\pmb{\uparrow}$})& SRCC ({\color{green!60!black}\bfseries$\pmb{\uparrow}$}) \\
\midrule
rFID$^{\dagger}$ & -0.04 & -0.31 & -0.07 & -0.31 & -0.06 & -0.21 & -0.15 & -0.31 & -- & -- & -- & -- \\
iFID$^{\dagger}$ & 0.85 & 0.86 & 0.82 & 0.84 & 0.89 & 0.91 & 0.88 & 0.92 & -- & -- & -- & -- \\
rFID$^{\star}$ & -0.11 & -0.36 & -0.11 & -0.30 & -0.13 & -0.34 & -0.15 & -0.39 & -0.10 & -0.30 & -0.10 & -0.28 \\
iFID$^{\star}$ & 0.86 & 0.83 & 0.80 & 0.79 & 0.90 & 0.88 & 0.88 & 0.90 & 0.72 & 0.77 & 0.64 & 0.70 \\
GAR-FID ($\eta_t=0.6$) & 0.81 & 0.79 & 0.77 & 0.78 & 0.79 & 0.84 & 0.80 & 0.79 & 0.86 & 0.89 & 0.84 & 0.85 \\
GAR-FID ($\eta_t=0.7$) & 0.93 & 0.90 & 0.89 & 0.87 & 0.91 & 0.90 & 0.91 & 0.86 & 0.95 & 0.96 & 0.93 & 0.91 \\
GAR-FID ($\eta_t=0.8$) & 0.98 & 0.96 & 0.96 & 0.94 & 0.98 & 0.99 & 0.97 & 0.96 & 0.99 & 0.99 & 0.97 & 0.95 \\
GAR-FID ($\eta_t=0.9$) & \textbf{1.00} & \textbf{0.99} & \textbf{0.98} & 0.98 & \textbf{1.00} & \textbf{1.00} & \textbf{0.98} & \textbf{0.96} & \textbf{1.00} & \textbf{1.00} & \textbf{0.98} & \textbf{0.97} \\
GAR-FID ($\eta_t=1.0$) & \textbf{1.00} & \textbf{0.99} & \textbf{0.98} & \textbf{0.99} & \textbf{1.00} & \textbf{1.00} & \textbf{0.98} & \textbf{0.96} & \textbf{1.00} & \textbf{1.00} & \textbf{0.98} & \textbf{0.97} \\
\bottomrule
\end{tabular}
}
\vspace{-2ex}
\end{table}
As shown in Table~\ref{tab:resultfid}, rFID exhibits negative correlation with gFID across SiT-B, SiT-XL, and mixed-generator configurations, consistent with the reconstruction-generation discrepancy observed in prior work~\citep{Yao2025ReconstructionVG,Kouzelis2025EQVAEER,Ye2025DistributionMV,Skorokhodov2025ImprovingTD,Chen2025MaskedAA,Xu2026MakingRF}. By contrast, the correlation between GAR-FID and gFID strengthens as $\eta_t$ increases. At $\eta_t=0.8$, GAR-FID already exhibits consistently high correlation across SiT-B, SiT-XL, and their mixed configurations, with SRCC ranging from $0.94$ to $0.99$.

The mixed SiT-B/XL setting further tests whether the GAR-FID-gFID relationship remains stable across generator scales. Correlations computed within a single generative model, such as SiT-B or SiT-XL~\citep{Ma2024SiTEF}, primarily capture tokenizer variation under a fixed generator scale. By contrast, the mixed setting jointly varies tokenizer choice and generator scale, providing a more demanding test that better reflects cross-system comparisons in latent generative modeling. In this setting, the correlation of iFID with gFID decreases, whereas GAR-FID maintains consistently strong correlation; for example, at $\eta_t=0.8$, GAR-FID achieves SRCCs of $0.99$ and $0.95$ without and with CFG, respectively, compared with $0.77$ and $0.70$ for iFID. These results indicate that the GAR-FID--gFID relationship remains robust across heterogeneous tokenizer-generator configurations.

\paragraph{Leave-one-out robustness across tokenizers.}
To test whether the GAR-FID--gFID correlation is driven by any individual tokenizer, we perform a leave-one-out analysis over the 14 SiT-XL tokenizer configurations, recomputing SRCC after omitting each tokenizer. Correlations are more sensitive to individual tokenizers at low noise levels but become uniformly high at $\eta_t\geq0.8$. At $\eta_t=0.8$, the leave-one-out SRCC is $0.99$ for every omission without CFG and ranges from $0.96$ to $0.97$ with CFG, showing that the strong correlation is not driven by any single tokenizer. Full results are provided in Tables~\ref{tab:loo-srcc-nocfg} and~\ref{tab:loo-srcc-cfg} in Appendix~\ref{appendix:additional_generalization_results}.

\paragraph{Computational cost.} For multi-step generators, intermediate GAR-FID can additionally be more computationally efficient than endpoint gFID, as it requires only a partial generative trajectory. Detailed runtime comparisons are provided in Table~\ref{tab:gar_runtime} in Appendix~\ref{appendix:gar_runtime}.

\section{A Unified Perspective on Latent Distribution Mismatch}
\label{sec:discussion}

As discussed in Section~\ref{sec:rfid_gfid}, latent distribution mismatch provides a useful explanation for the gap between reconstruction and generation: the same decoder operates under different latent distributions in the two regimes. GAR makes this shift observable by exposing decoder behavior along the intermediate distributions $\mathcal{P}_g^t$ between $\mathcal{P}_e$ and $\mathcal{P}_g$. Beyond this diagnostic role, GAR reveals a broader insight: \emph{generation-oriented evaluation and training should account for the latent distributions encountered during generation}.

This perspective naturally leads to two complementary ways to address the latent distribution mismatch. The first is to improve the latent generative process, such that the resulting distribution $\mathcal{P}_g$ more faithfully approximates the encoder-induced distribution $\mathcal{P}_e$ that the generator is trained to model. The second is to adapt the decoder, such that it reliably operates on samples drawn from $\mathcal{P}_g$ or intermediate generative distributions $\mathcal{P}_g^t$. These correspond to reducing the mismatch on the generator side, or adapting the decoder to the latent distributions encountered during generation.

A large body of prior work has focused on the first direction, improving the latent generative process through multiple complementary avenues, including stronger architectures and scaling strategies, improved objectives and transport formulations, enhanced multimodal conditioning, and efficient sampling or distillation techniques~\citep{Peebles2022ScalableDM,esser2024scaling,Mei2024BiggerIN,Tong2026ScalingTD,Chen2023PixArtFT,Lipman2023FlowMF,Liu2023FlowSA,Albergo2023BuildingNF,Geng2025MeanFF,wu2025qwen,flux2024,Xie2025SANAEH,ProgressiveDistillation2022,Song2023ConsistencyM,Luo2023LatentCM,Song2023ImprovedTF,Sauer2024FastHI}.
Additional approaches leverage reinforcement learning, preference-based optimization, or richer latent representations to further improve generative quality~\citep{Yu2024RepresentationAF,Liu2025FlowGRPOTF,Xue2025DanceGRPOUG,Lee2023AligningTM,Fan2023DPOKRL,Xu2023ImageRewardLA,Wallace2023DiffusionMA,zheng2026diffusion,Leng2025REPAEUV,yang2026latent,Chen2025MaskedAA,Chen2024SoftVQVAEE1,Ye2025DistributionMV}. Although these methods do not necessarily explicitly optimize the discrepancy between $\mathcal{P}_e$ and $\mathcal{P}_g$, they broadly improve the latent representation being modeled. In the standard two-stage setting, where the latent generator is trained to model the encoder-induced distribution $\mathcal{P}_e$, improving its modeling fidelity can be viewed as reducing the generator-side mismatch between $\mathcal{P}_g$ and $\mathcal{P}_e$. Consistent with this perspective, our controlled study shows that stronger generators exhibit a smaller measured discrepancy between $\mathcal{P}_e$ and $\mathcal{P}_g$.

Our results motivate the complementary decoder-side perspective. Even as the generative model improves, generation still relies on a decoder trained under $\mathcal{P}_e$ but operating on latents drawn from $\mathcal{P}_g$. Improving the generator therefore does not directly optimize decoder behavior under generation-time latent distributions. Moreover, as evidenced by the \emph{decoder amplification effect}, discrepancies in latent space can become substantially more pronounced in downstream decoder representations. Consequently, generative performance depends not only on the mismatch between $\mathcal{P}_e$ and $\mathcal{P}_g$, but also on how reliably $\mathcal{D}_{\phi}$ operates as its input distribution shifts toward the generation-time regime.

This motivates aligning decoder training with generation-relevant latent distributions. Intermediate GAR latents provide a practical way to do so: they move toward the generation-time regime while retaining correspondence with their source images, thereby preserving paired supervision that is unavailable for fully generated latents. We instantiate this idea through \emph{decoder adaptation} (Section~\ref{sec:method}), which fine-tunes the decoder on intermediate GAR latents while keeping the encoder and generative model fixed. In this way, the same GAR trajectory used to diagnose the reconstruction-generation shift also provides a natural training distribution for improving downstream generative performance.

\section{Decoder Adaptation via Latent Distribution Alignment}
\label{sec:method}
We instantiate the decoder-side perspective developed above through \emph{decoder adaptation} (DA), which keeps the encoder $\mathcal{E}_{\theta}$ and generative model $\mathcal{G}_{\varphi}$ fixed and updates only the decoder $\mathcal{D}_{\phi}$ on intermediate GAR latents. Given an image $x$, we first obtain its encoder latent $z_e=\mathcal{E}_{\theta}(x)\sim\mathcal{P}_e$. Following Section~\ref{sec:GAR}, we perturb $z_e$ as $z^t_{\mathrm{noisy}}=a_t z_e+b_t\epsilon$, where $\epsilon\sim\mathcal{N}(0,I)$, and denoise it through the fixed generative model to obtain $z_g^t=\mathcal{G}_{\varphi}(z^t_{\mathrm{noisy}},t)\sim\mathcal{P}_g^t$, which is decoded as $\bar{x}(\phi)=\mathcal{D}_{\phi}(z_g^t)$. Only the decoder parameters $\phi$ are optimized during adaptation.

The choice of intermediate GAR latents is central to DA. Standard decoder training operates on encoder latents $z_e\sim\mathcal{P}_e$, which retain source correspondence but do not expose the decoder to generation-relevant latent distributions. Fully generated latents $z_g\sim\mathcal{P}_g$, in contrast, reflect the generation-time distribution but have no paired target images. Intermediate GAR latents bridge these regimes: as $\eta_t$ increases, $\mathcal{P}_g^t$ moves toward $\mathcal{P}_g$, while intermediate noise levels retain correspondence with the source image $x$. DA therefore trains the decoder on generation-relevant latents while preserving paired supervision. In this sense, latent distribution alignment is achieved by changing the decoder's training distribution rather than introducing an explicit distribution-matching objective. We use noise levels $\eta_t\in[0.25,0.45]$ for adaptation, with sensitivity studied in Appendix~\ref{appendix:noise_level_choice}.

To optimize the decoder, we adopt standard objectives commonly used in modern visual tokenizers~\citep{Tian2024VisualAM,Fang2025VQTransplant,Yu2026AutoregressiveIG}, combining pixel-level, perceptual, adversarial, and feature-statistics supervision:
\begin{equation}
    \mathcal{L}_{\text{DA}}(\phi) = 
    \lambda_1 \| \bar{x}(\phi) - x \|_2^2 
    + \lambda_2 \mathcal{L}_{\text{LPIPS}} 
    + \lambda_3 \mathcal{L}_{\text{GAN}} 
    + \lambda_4 \mathcal{L}_{\text{Gram}}.
\end{equation}
The pixel, LPIPS, and Gram terms exploit the correspondence between the GAR output $\bar{x}(\phi)$ and its source image $x$. Specifically, $\mathcal{L}_{\text{LPIPS}}$ measures perceptual similarity using deep feature representations~\citep{Zhang2018TheUE,Simonyan2014VeryDC}, while $\mathcal{L}_{\text{Gram}}$ matches feature covariance statistics through Gram matrices~\citep{Gatys2016ImageST,Lu2025ATokenAU}. The adversarial term $\mathcal{L}_{\text{GAN}}$ encourages realistic decoded outputs and is implemented using a hinge-based adversarial objective~\citep{Isola2016ImagetoImageTW,Lim2017GeometricG,karras2019style,Karras2019AnalyzingAI}. We set $\lambda_1=1.0$, $\lambda_2=1.0$, $\lambda_3=0.7$, and $\lambda_4=10.0$ across all experiments. Importantly, these objectives are standard; the key change in DA is the training distribution presented to the decoder, replacing encoder-only latents with paired, generation-aware GAR latents.


\section{Experiments}
\label{sec:experiments}
\vspace{-1ex}
\paragraph{Experimental Setup.}
To evaluate decoder adaptation, we conduct experiments on four iMF model scales, iMF-B/2, iMF-M/2, iMF-L/2, and iMF-XL/2~\citep{Geng2025ImprovedMF}, all sharing the SD-VAE tokenizer~\citep{Rombach2022HighResolutionIS}. During adaptation, the tokenizer encoder and latent generative model are frozen, and only the decoder is updated. We consider two settings: DA w/o CFG, which uses GAR latents generated without CFG and trains for 10 epochs, and CFG-aware DA, which uses CFG during GAR construction and trains for one epoch. Experiments are conducted on ImageNet-1k at $256 \times 256$ resolution~\citep{Deng2009ImageNetAL}. Additional implementation details are provided in Appendix~\ref{app:exp_details}. Code and checkpoints will be publicly released.

\begin{table}[!t]
\vspace{-1ex}
\centering
\caption{\small{\textbf{Effect of decoder adaptation (DA) on generative performance.} We report gFID without and with CFG across iMF model scales under two adaptation settings: DA w/o CFG, trained for 10 epochs using GAR latents generated without CFG, and CFG-aware DA, trained for one epoch using GAR latents generated with CFG.
$^{\dagger}$: results reported in iMF~\citep{Geng2025ImprovedMF}.
$^{\ddagger}$: our reproduction using the official iMF protocol.
$^{\star}$: our reproduction using the OpenAI protocol.
See Appendix~\ref{appendix:gfid_protocol} for protocol details.}}
\vspace{-2ex}
\label{tab:generative_performance}
\resizebox{\linewidth}{!}{
\begin{tabular}{@{}lcccccccc@{}}
\toprule
\multirow{2}{*}{Setting} & \multicolumn{4}{c}{gFID w/o CFG ({\color{green!60!black}\bfseries$\pmb{\downarrow}$})} & \multicolumn{4}{c}{gFID w/ CFG ({\color{green!60!black}\bfseries$\pmb{\downarrow}$})}\\
\cmidrule(lr){2-5} \cmidrule(lr){6-9}
& iMF-B/2 & iMF-M/2 & iMF-L/2 & iMF-XL/2 & iMF-B/2 & iMF-M/2 & iMF-L/2 & iMF-XL/2 \\
\midrule
w/o DA$^{\dagger}$ & -\quad\quad\enspace & -\quad\quad\enspace & -\quad\quad\enspace & -\quad\quad\enspace & 3.39\quad\quad\enspace & 2.27\quad\quad\enspace & 1.86\quad\quad\enspace & 1.72\quad\quad\enspace \\
\midrule
w/o DA$^{\ddagger}$ & 16.58\quad\quad\enspace & 11.92\quad\quad\enspace & 9.28\quad\quad\enspace & 9.78\quad\quad\enspace & 3.37\quad\quad\enspace & 2.27\quad\quad\enspace & 1.86\quad\quad\enspace & 1.73\quad\quad\enspace \\
DA w/o CFG (10 ep.)$^{\ddagger}$ & 12.21 { \color{red} $\uparrow _{4.37}$} & 9.52 { \color{red} $\uparrow _{2.40}$}& 7.34 { \color{red} $\uparrow _{1.94}$}& 7.47 { \color{red} $\uparrow _{2.31}$}& 2.90 { \color{red} $\uparrow _{0.47}$}& 2.12 { \color{red} $\uparrow _{0.15}$} & 1.65 { \color{red} $\uparrow _{0.21}$} & 1.56 { \color{red} $\uparrow _{0.17}$}\\
CFG-aware DA(1 ep.)$^{\ddagger}$ & -\quad\quad\enspace & -\quad\quad\enspace & -\quad\quad\enspace & -\quad\quad\enspace & 2.81 { \color{red} $\uparrow_{0.56}$} & 1.98 { \color{red} $\uparrow_{0.29}$} & 1.58 { \color{red} $\uparrow_{0.28}$} & 1.50 { \color{red} $\uparrow_{0.23}$}\\
\midrule
w/o DA$^{\star}$ & 16.41\quad\quad\enspace & 11.92\quad\quad\enspace & 9.26\quad\quad\enspace & 9.41\quad\quad\enspace & 3.47\quad\quad\enspace & 2.40\quad\quad\enspace & 1.90\quad\quad\enspace & 1.80\quad\quad\enspace \\
DA w/o CFG (10 ep.)$^{\star}$ & 12.41 { \color{red} $\uparrow _{4.00}$}& 9.66 { \color{red} $\uparrow _{2.26}$} & 7.41 { \color{red} $\uparrow _{1.85}$} & 7.61 { \color{red} $\uparrow _{1.80}$} & 3.12 { \color{red} $\uparrow _{0.35}$} & 2.30 { \color{red} $\uparrow _{0.10}$} & 1.80 { \color{red} $\uparrow _{0.10}$} & 1.73 { \color{red} $\uparrow _{0.07}$} \\
CFG-aware DA (1 ep.)$^{\star}$ & -\quad\quad\enspace & -\quad\quad\enspace & -\quad\quad\enspace & -\quad\quad\enspace & 3.01 { \color{red} $\uparrow_{0.46}$} & 2.19 { \color{red} $\uparrow_{0.21}$} & 1.70 { \color{red} $\uparrow_{0.20}$} & 1.61 { \color{red} $\uparrow_{0.19}$}\\
\toprule
\end{tabular}
}
\vspace{-2ex}
\end{table}

\paragraph{Main Results.}
As shown in Table~\ref{tab:generative_performance}, DA w/o CFG consistently improves generative performance across all model scales, with larger gFID gains observed without CFG and for smaller models. CFG-aware DA further improves guided generation despite using only one epoch of adaptation, outperforming or matching the 10-epoch DA w/o CFG setting across all reported gFID evaluations with CFG. The gains remain consistent across both the official iMF and OpenAI evaluation protocols, suggesting that the improvement is not tied to a particular FID implementation. 
Importantly, neither adaptation setting introduces additional inference cost, as only the decoder weights are updated while the inference pipeline and generative model remain unchanged.

\vspace{-1ex}
\paragraph{Additional Analyses.} We provide further analyses of the 10-epoch DA w/o CFG setting in Appendix~\ref{appendix:additional decoder adaptation}. First, repeated evaluation over three random seeds shows consistently small variance and stable gFID improvements across model scales (Table~\ref{tab:seed_variance}). Second, after adaptation, rFID varies more consistently with gFID across generator scales, indicating a closer correspondence between reconstruction and generation behavior (Figure~\ref{fig:FID-post-train}). Third, the noise-level sensitivity analysis in Table~\ref{tab:noise levels choices} reveals a clear trade-off between generation awareness and source correspondence, with $\eta_t=0.5$ achieving the best gFID without CFG across all scales under fixed noise. Fourth, reconstruction results in Table~\ref{tab:post_train_reconstruction} show that DA trades standard reconstruction fidelity for improved behavior as GAR approaches the generation-time regime. Finally, qualitative comparisons show that DA reduces visual artifacts and yields richer details and more semantically consistent samples (Figures~\ref{fig:appendix_da_class_014}-\ref{fig:appendix_da_class_817}).


\paragraph{Generalization Study.}
Beyond iMF~\citep{Geng2025ImprovedMF}, we further apply the same one-epoch CFG-aware decoder adaptation to MeanFlow-XL/2~\citep{Geng2025MeanFF} and DMF-XL/2+~\citep{Lee2025DecoupledMT}. Under the official iMF evaluation protocol, decoder adaptation consistently improves gFID from $3.43$ to $2.96$ on MeanFlow-XL/2 and from $2.16$ to $1.90$ on DMF-XL/2+. These results show that the benefits of decoder adaptation extend beyond iMF and remain effective across both from-scratch and distillation-based generative models, supporting its generalization across different generator training paradigms. A broader system-level comparison on class-conditional ImageNet $256\times256$ is provided in Table~\ref{tab:all_results} of Appendix~\ref{appendix:additional decoder adaptation}.

\section{Conclusion}
\label{sec:conclusion}

This work presents a latent distribution perspective on the reconstruction-generation discrepancy in latent generative models. We identify that reconstruction and generation evaluate the same decoder under fundamentally different latent distributions, and introduce GAR to make this shift observable through a controlled trajectory. GAR-FID provides a reliable proxy for generative performance, and decoder adaptation on intermediate GAR latents consistently improves generation quality across scales. We hope this perspective encourages future work to account for generation-time latent distributions in tokenizer evaluation and training.

\printbibliography

\clearpage

\begin{appendices}
\crefalias{section}{appsec}
\crefalias{subsection}{appsec}
\crefalias{subsubsection}{appsec}

\setcounter{equation}{0}
\renewcommand{\theequation}{\thesection.\arabic{equation}}

\Huge{Appendix}

\small

\section{Latent Fréchet Distance}
\label{appendix:latent-fd}

To measure the distributional discrepancy in latent space, we adopt the Fréchet distance between two sets of latent variables, $z_i^{a} \sim \mathcal{P}_A$ and $z_i^{b} \sim \mathcal{P}_B$. Specifically, we estimate the empirical mean and covariance of the two latent distributions, denoted by $(\mu_1, \Sigma_1)$ and $(\mu_2, \Sigma_2)$, respectively. Based on these statistics, we define the latent Fréchet distance (Latent-FD) as
$
\mathrm{FD}(\mathcal{P}_A, \mathcal{P}_B) 
= \|\mu_1 - \mu_2\|_2^2 
+ \mathrm{Tr}\!( \Sigma_1 + \Sigma_2 - 2(\Sigma_1 \Sigma_2)^{1/2} ).
$
This metric provides a well-established measure of discrepancy between two distributions by capturing both first-order (mean) and second-order (covariance) statistics, and is closely related to the widely used Fréchet Inception Distance (FID). Both metrics compute the Fréchet distance under a Gaussian assumption, differing only in the representation space used to estimate these statistics. Specifically, FID operates in a learned semantic feature space, where each image is mapped to a 2048-dimensional representation using a pretrained InceptionV3~\citep{Szegedy2016RethinkingTI} network. In contrast, our Latent-FD is computed directly in the model latent space, using latents from the encoder or generative process. This distinction allows Latent-FD to directly quantify distributional discrepancies relevant to the generative pipeline, while retaining the same underlying geometric interpretation as FID.

\section{Proof of Proposition~\ref{prop:latent-fd-gfid-bound}}
\label{app:proof of proposition}
Let $\mathcal{P}_{\mathrm{data}}$ denote the real image distribution, let $\psi$ be the Inception feature extractor used for FID, and define $T=\psi\circ\mathcal{D}_{\phi}$. We write
\[
\mathcal{Q}_{\mathrm{data}}=\psi_{\#}\mathcal{P}_{\mathrm{data}},\qquad
\mathcal{Q}_{e}=T_{\#}\mathcal{P}_{e},\qquad
\mathcal{Q}_{g}=T_{\#}\mathcal{P}_{g}.
\]
For any distribution $\mathcal{Q}$ with finite second moment, let $\Gamma(\mathcal{Q})=\mathcal{N}(\mu_{\mathcal{Q}},\Sigma_{\mathcal{Q}})$ be the Gaussian distribution with the same mean and covariance as $\mathcal{Q}$. The FID between two feature distributions $\mathcal{Q}$ and $\mathcal{R}$ is the squared 2-Wasserstein distance between their Gaussian approximations:
\[
\mathrm{FID}(\mathcal{Q},\mathcal{R})=
W_2^2\!\left(\Gamma(\mathcal{Q}),\Gamma(\mathcal{R})\right).
\]
Therefore, using the triangle inequality for $W_2$,
\[
\begin{aligned}
\sqrt{\mathrm{gFID}}
&=W_2\!\left(\Gamma(\mathcal{Q}_{\mathrm{data}}),\Gamma(\mathcal{Q}_{g})\right)\\
&\leq
W_2\!\left(\Gamma(\mathcal{Q}_{\mathrm{data}}),\Gamma(\mathcal{Q}_{e})\right)
+
W_2\!\left(\Gamma(\mathcal{Q}_{e}),\Gamma(\mathcal{Q}_{g})\right)\\
&=
\sqrt{\mathrm{rFID}}
+
\sqrt{\mathrm{FID}(\mathcal{Q}_{e},\mathcal{Q}_{g})}.
\end{aligned}
\]
The Gaussian Fréchet distance between the moment-matched Gaussians is upper bounded by the true Wasserstein distance between the original distributions, since
\[
\mathrm{FID}(\mathcal{Q}_{e},\mathcal{Q}_{g})
\leq
W_2^2(\mathcal{Q}_{e},\mathcal{Q}_{g}).
\]
Moreover, because $T$ is $L$-Lipschitz, pushing any coupling of $\mathcal{P}_e$ and $\mathcal{P}_g$ through $T$ gives
\[
W_2(\mathcal{Q}_{e},\mathcal{Q}_{g})
=
W_2(T_{\#}\mathcal{P}_{e},T_{\#}\mathcal{P}_{g})
\leq
L\,W_2(\mathcal{P}_{e},\mathcal{P}_{g}).
\]
Finally, since $\mathcal{P}_e$ and $\mathcal{P}_g$ are Gaussian, the latent Fréchet distance equals the squared Wasserstein distance between them:
\[
\mathrm{Latent\text{-}FD}(\mathcal{P}_e,\mathcal{P}_g)
=
W_2^2(\mathcal{P}_e,\mathcal{P}_g).
\]
Combining the preceding inequalities yields
\[
\sqrt{\mathrm{gFID}}
\leq
\sqrt{\mathrm{rFID}}
+
L_{\psi}L_{\mathcal{D}}\sqrt{\mathrm{Latent\text{-}FD}(\mathcal{P}_e,\mathcal{P}_g)}.
\]

\section{Structural and Perceptual Quality under GAR}
\label{appendix:structural-perceptual}

We provide additional analysis of reconstruction quality under GAR using both structural and perceptual metrics. This analysis complements the main text by illustrating how latent distribution shift manifests in reconstruction quality. For structural fidelity, we consider peak signal-to-noise ratio (PSNR) and structural similarity index (SSIM). 
For perceptual quality, we use Fréchet Inception Distance~\citep[FID;][]{Heusel2017GANsTB} and learned perceptual image patch similarity~\citep[LPIPS;][]{Zhang2018TheUE}. 
When computed under the GAR framework, we refer to these metrics as GAR-PSNR, GAR-SSIM, GAR-FID, and GAR-LPIPS, respectively. All experiments are conducted on the ImageNet validation set~\citep{Deng2009ImageNetAL}, using the SD-VAE tokenizer~\citep{Rombach2022HighResolutionIS} and two representative flow-matching models, iMF-B/2 and iMF-XL/2~\citep{Geng2025ImprovedMF}.

\begin{figure*}[!t]
    \centering
    \includegraphics[width=1.0\linewidth]{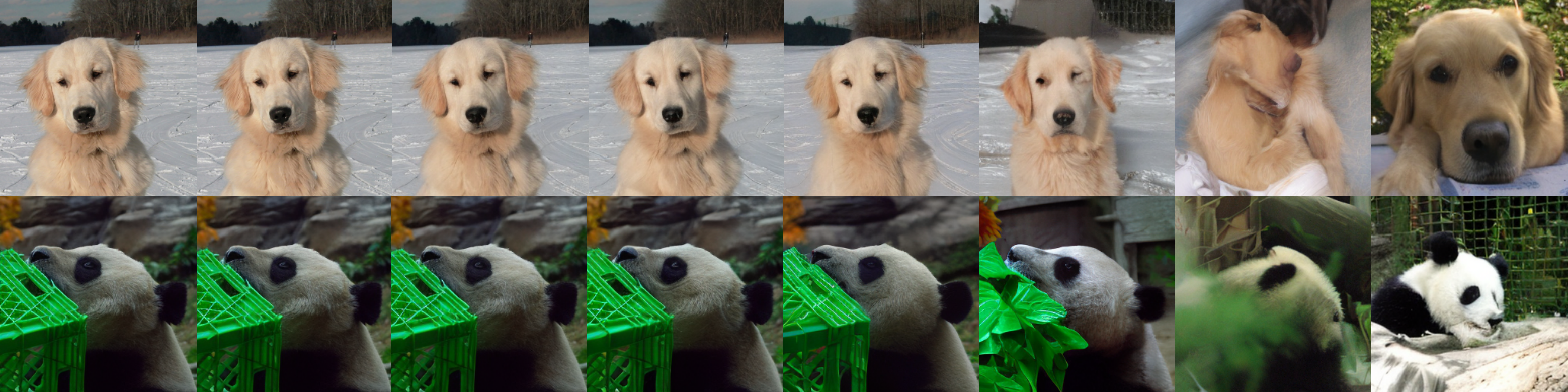}
    \vspace{-4ex}
    \caption{\small{Illustration of the progressive loss of source structure along the GAR trajectory. Each row shows an original input image (leftmost column) followed by GAR reconstructions at noise levels $\eta_t \in \{0.0, 0.1, 0.2, 0.4, 0.6, 0.8, 1.0\}$ from left to right. As the noise level increases, reconstructions progressively deviate from the original inputs: while coarse semantic content may remain loosely recognizable at intermediate noise levels, fine-grained structural details and local textures are gradually degraded. At high noise levels (e.g., $\eta_t \geq 0.6$), the reconstructed images no longer preserve the original structure, resulting in semantically plausible but structurally inconsistent outputs. This progression illustrates how source-level structural fidelity weakens as GAR moves toward the generation endpoint.}}
    \label{fig:GAR-examples}
    \vspace{-1ex}
\end{figure*}

\begin{figure}[!t]
\centering
\small
\vspace{-1mm}
    \begin{subfigure}{0.495\linewidth}
        \includegraphics[width=\linewidth]{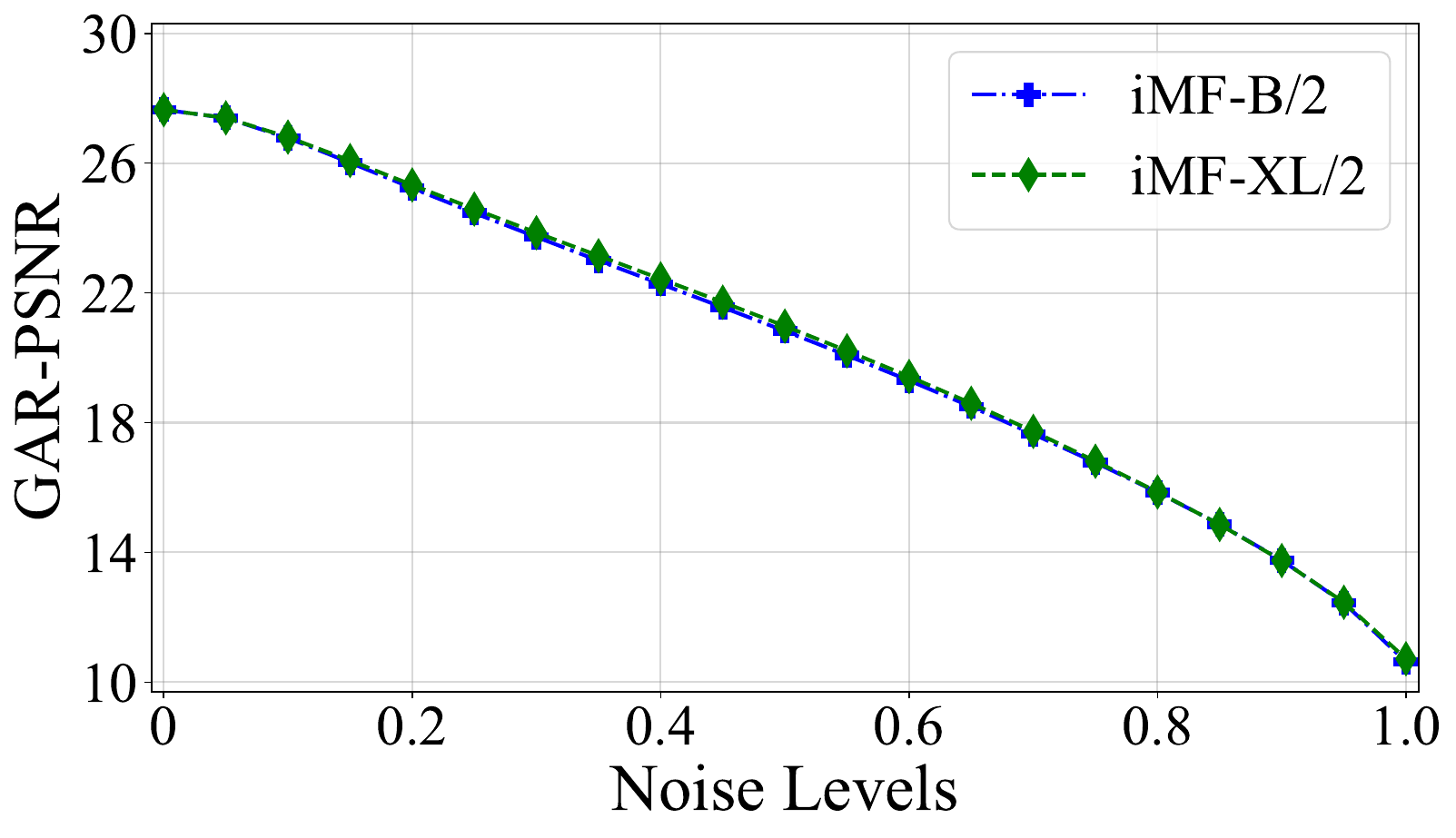}
        \vspace{-5mm}
        \caption{PSNR vs. noise level.}
    \end{subfigure}
    \hfill
    \begin{subfigure}{0.495\linewidth}
        \includegraphics[width=\linewidth]{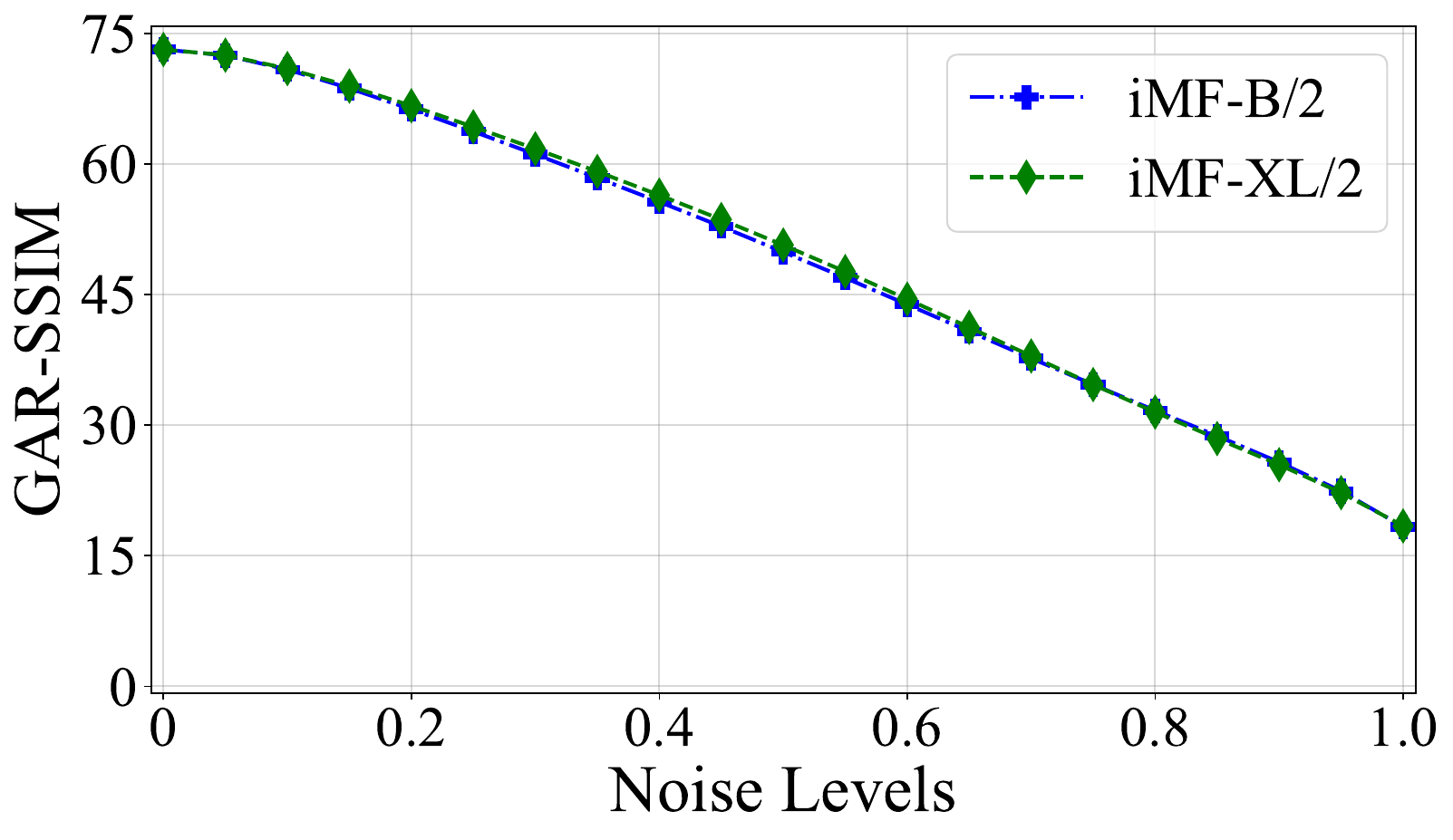}
        \vspace{-5mm}
        \caption{SSIM vs. noise level.}  
    \end{subfigure}
\vspace{-7mm}
\caption{\small{Structural quality under GAR across noise levels. PSNR (left) and SSIM (right) degrade consistently as noise increases. Increasing generative model capacity (iMF-B/2 vs. iMF-XL/2) does not improve structural quality, as both follow nearly identical trends.}}
\label{fig:structural_quality}
\vspace{-1mm}
\end{figure}

\begin{figure}[!t]
\centering
\vspace{-2mm}
    \begin{subfigure}{0.495\linewidth}
        \includegraphics[width=\linewidth]{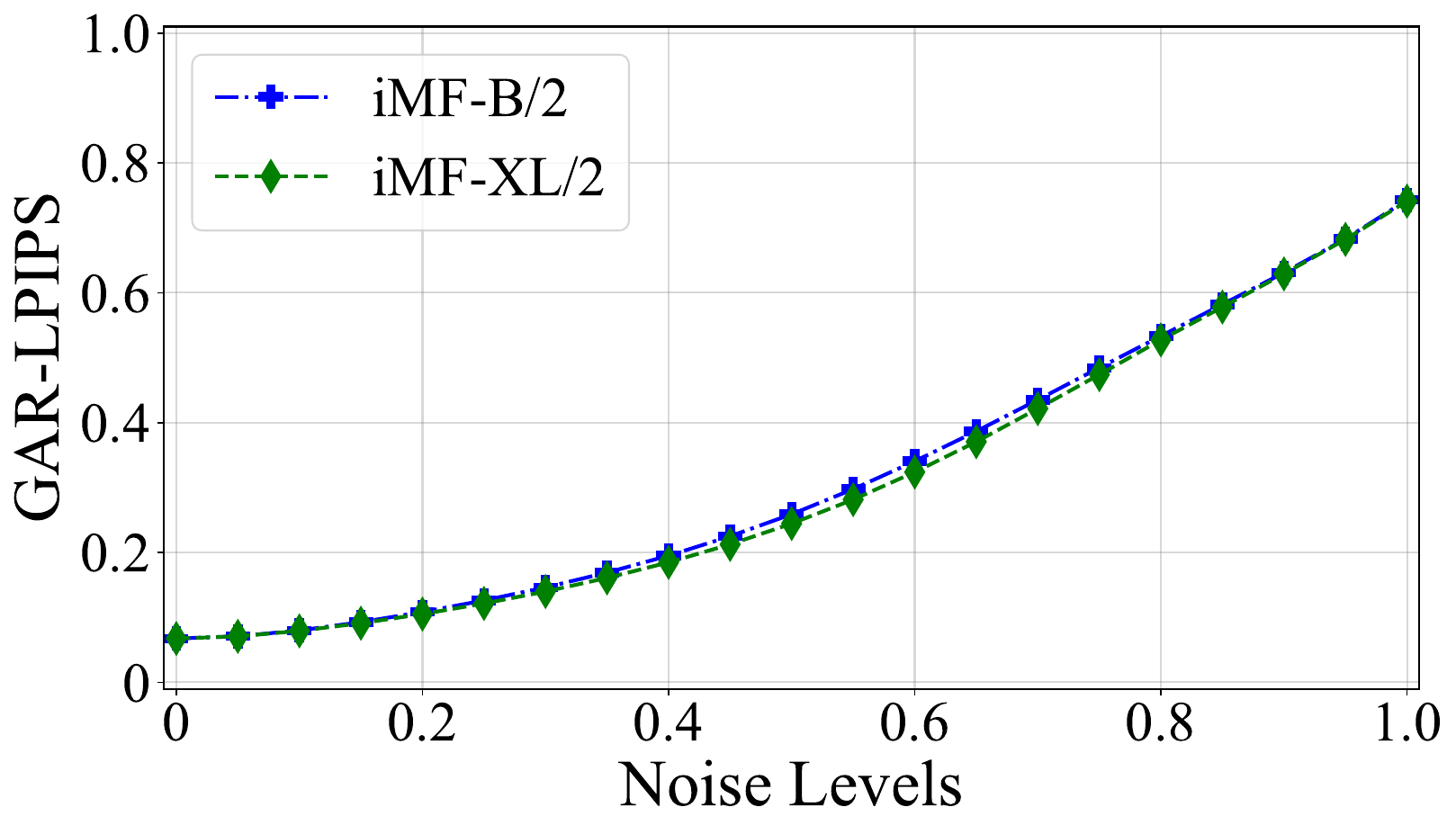}
        \vspace{-5mm}
        \caption{LPIPS vs. noise level.}
    \end{subfigure}
    \hfill
    \begin{subfigure}{0.495\linewidth}
        \includegraphics[width=\linewidth]{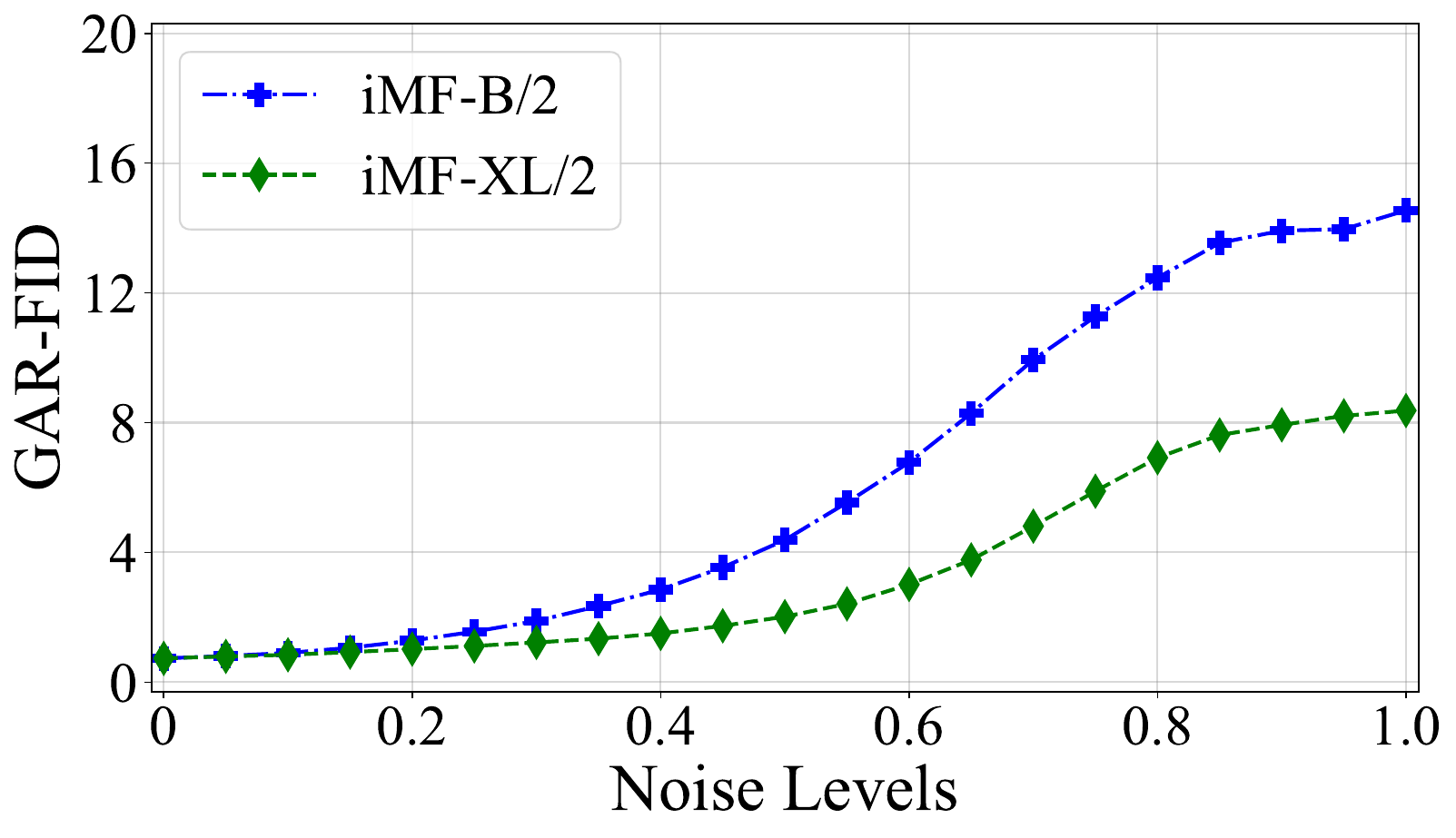}
        \vspace{-5mm}
        \caption{FID vs. noise level.}
    \end{subfigure}
\vspace{-7mm}
\caption{\small{Perceptual quality under GAR across noise levels. LPIPS (left) and FID without CFG (right) both degrade as noise increases. Increasing model capacity (iMF-B/2 vs. iMF-XL/2) has little effect on LPIPS but significantly improves FID across noise levels.}}
\label{fig:perceptual_quality}
\vspace{-2mm}
\end{figure}

\paragraph{Structural Quality.}
As depicted in Figure~\ref{fig:structural_quality}, both structural metrics, PSNR and SSIM, decline consistently as the noise level increases. At sufficiently large noise levels (e.g., $\eta_t \geq 0.6$), GAR outputs deviate substantially from the original inputs: coarse semantic content may remain partially recognizable, while fine-grained structure and local textures are increasingly altered, as illustrated in Figure~\ref{fig:GAR-examples}. This behavior is expected as GAR moves toward the generation endpoint and source correspondence progressively weakens. Increasing generative model capacity (iMF-B/2 vs.\ iMF-XL/2) has little effect on these structural metrics, with both models exhibiting nearly identical trends across noise levels.

\paragraph{Perceptual Quality.}
As shown in Figure~\ref{fig:perceptual_quality}, both perceptual metrics, LPIPS and FID without CFG, increase consistently as the noise level grows (higher values indicate worse quality). Notably, increasing generative model capacity (iMF-B/2 vs.\ iMF-XL/2) has little effect on LPIPS, as the curves nearly overlap, but leads to a substantial improvement in FID across all noise levels. This indicates that increasing generative model capacity has little effect on input-level perceptual similarity as measured by LPIPS, while yielding better distribution-level FID.

\paragraph{Latent Distribution Shift Perspective.} The trends above are consistent with the latent distribution shift induced along the GAR trajectory. As shown in Figure~\ref{fig:latent-fd}, increasing the noise level moves $\mathcal{P}_g^t$ progressively away from $\mathcal{P}_e$ and toward $\mathcal{P}_g$. At the same time, source correspondence weakens, leading GAR outputs to become less similar to their corresponding input images. This behavior is reflected by the decreasing PSNR and SSIM and the increasing LPIPS and FID observed along the trajectory. Together, these results illustrate how the transition from reconstruction toward generation is accompanied by a progressive loss of source-level structural and perceptual fidelity.

\section{Intermediate Latent-FD as an Indicator of Final Latent Mismatch}
\label{appendix:intermediate_latent_fd}

\begin{figure}[h]
\centering
\small
\vspace{-2mm}
\includegraphics[width=1.0\textwidth]{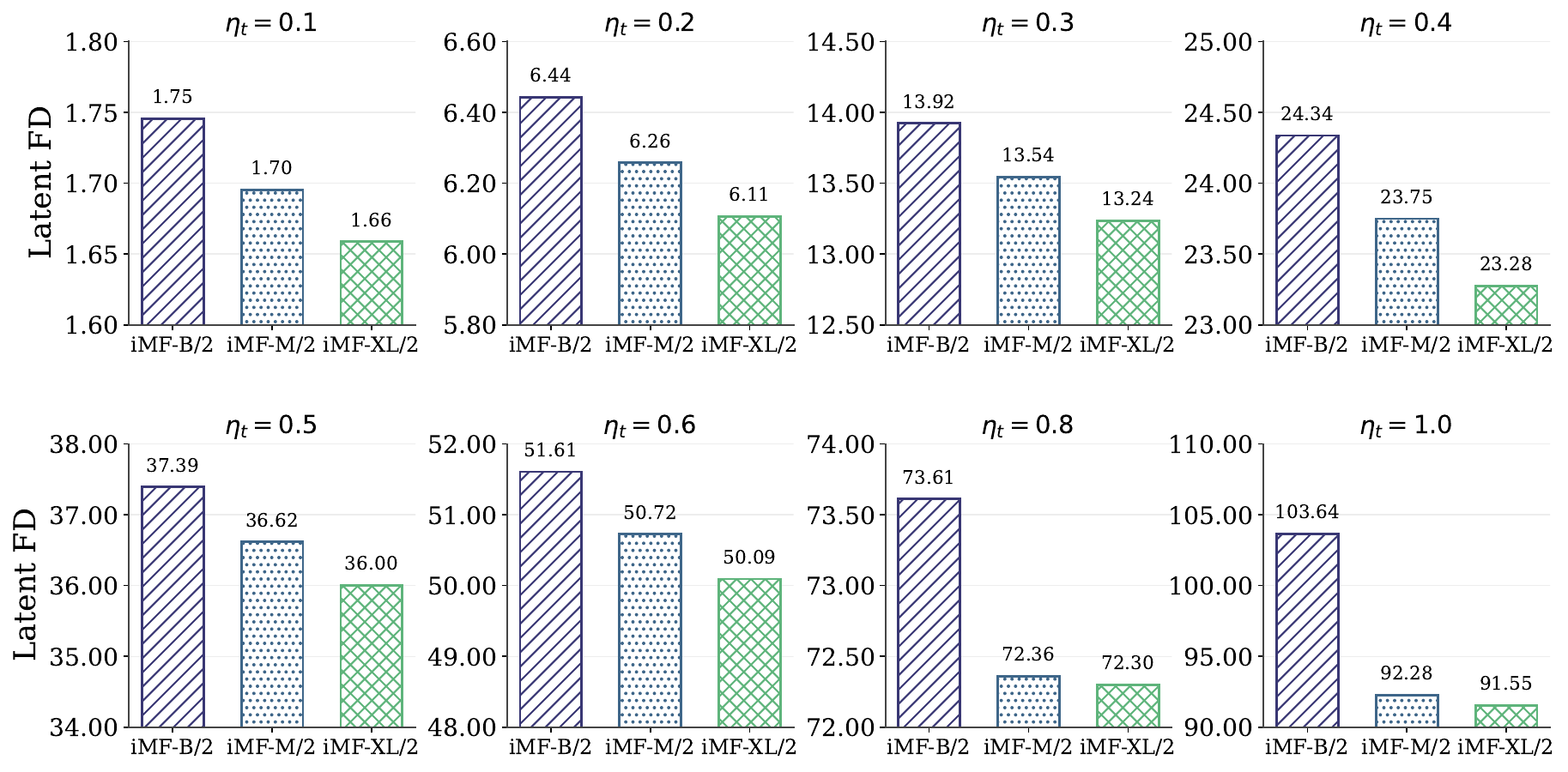}
\vspace{-7mm}
\caption{\small{
Latent-FD$(\mathcal{P}_g^t, \mathcal{P}_e)$ across noise levels for different generative models, with $\eta_t = 1.0$ corresponding to Latent-FD$(\mathcal{P}_g, \mathcal{P}_e)$. 
The model ordering is preserved across noise levels.
}}
\label{fig:latent-FD relationship}
\vspace{-2mm}
\end{figure}
Figure~\ref{fig:latent-FD relationship} compares $\mathrm{Latent\text{-}FD}(\mathcal{P}_g^t, \mathcal{P}_e)$ across noise levels for different generative models, where $\eta_t = 1.0$ corresponds to $\mathrm{Latent\text{-}FD}(\mathcal{P}_g, \mathcal{P}_e)$. 
We observe that the model ordering is preserved across all evaluated noise levels. This consistent ordering indicates that larger intermediate latent mismatch is associated with larger final mismatch at $\eta_t = 1.0$. 
We summarize this empirical relationship as
\[
\mathrm{Latent\text{-}FD}(\mathcal{P}_g^t, \mathcal{P}_e) \uparrow
\;\leadsto\;
\mathrm{Latent\text{-}FD}(\mathcal{P}_g, \mathcal{P}_e) \uparrow.
\]
Under the fixed-tokenizer setting, this ordering consistency supports using intermediate latent mismatch to characterize the relative final mismatch across generative models.

\section{Noise Level Choice in Decoder Adaptation}
\label{appendix:noise_level_choice}

A key practical consideration in decoder adaptation is the choice of noise level. Intermediate GAR latents can in principle be constructed using any $\eta_t\in(0,1)$, but the noise level controls the trade-off between generation awareness and source correspondence. At large noise levels, GAR latents move closer to the generation-time regime while progressively losing correspondence with the source image $x$, making paired reconstruction supervision less reliable. At small noise levels, the latents remain close to the encoder distribution $\mathcal{P}_e$, providing limited exposure to the latent distributions encountered during generation. We therefore sample intermediate noise levels as $\eta_t\sim\mathrm{Unif}[\eta_{\min},\eta_{\max}]$ to balance these two effects. In our experiments, we use $\eta_{\min}=0.25$ and $\eta_{\max}=0.45$, which provides effective decoder adaptation across the evaluated model scales.

\section{Computational Cost of rFID, GAR-FID, and gFID}
\label{appendix:gar_runtime}

\begin{wraptable}[21]{r}{0.52\textwidth}
\centering
\small
\vspace{-5.5mm}
\caption{\small{Computational cost of rFID, GAR-FID, and gFID on REPAE-VAE. rFID requires no generative-model evaluations, GAR-FID traverses a partial generation trajectory determined by $\eta_t$, and gFID uses the full 250-step generation process.}}
\label{tab:gar_runtime}
\vspace{-3mm}
\resizebox{0.52\textwidth}{!}{
\begin{tabular}{@{}lccc@{}}
\toprule
Metric & Noise Level $\eta_t$ & NFE & Time (hours) \\
\midrule
rFID & 0.0 & 0 & 0.095 \\
GAR-FID & 0.1 & 32 & 0.799 \\
GAR-FID & 0.2 & 56 & 1.327 \\
GAR-FID & 0.3 & 81 & 1.879 \\
GAR-FID & 0.4 & 105 & 2.407 \\
GAR-FID & 0.5 & 129 & 2.936 \\
GAR-FID & 0.6 & 153 & 3.465 \\
GAR-FID & 0.7 & 177 & 3.992 \\
GAR-FID & 0.8 & 202 & 4.545 \\
GAR-FID & 0.9 & 226 & 5.075 \\
GAR-FID & 1.0 & 250 & 5.572 \\
gFID & 1.0 & 250 & \textbf{5.570} \\
\bottomrule
\end{tabular}
}
\vspace{-5mm}
\end{wraptable}
We compare the computational cost of rFID, GAR-FID, and gFID using REPAE-VAE, whose full generation process requires 250 function evaluations (NFEs). The three metrics correspond to different positions along the reconstruction-generation trajectory. rFID evaluates the reconstruction endpoint and requires no generative-model evaluation, whereas gFID evaluates the generation endpoint and requires the full 250-step generative process. GAR-FID operates between these two endpoints: at noise level $\eta_t$, it starts from a partially noised encoder latent and applies only the remaining generative trajectory before decoding. Its computational cost therefore increases with $\eta_t$ and approaches that of gFID as $\eta_t\rightarrow1$. As shown in Table~\ref{tab:gar_runtime}, rFID is the least expensive, requiring only $0.095$ hours in this setting. GAR-FID exhibits a gradual increase in computational cost along the trajectory. For example, GAR-FID requires $2.936$ hours at $\eta_t=0.5$ and $4.545$ hours at $\eta_t=0.8$, corresponding to approximately $53\%$ and $82\%$ of the runtime of gFID, respectively. At $\eta_t=1.0$, GAR reaches the generation endpoint and its runtime ($5.572$ hours) is effectively identical to that of gFID ($5.570$ hours).

\section{Limitations}
\label{appendix:limitations}

Our work has several limitations.

First, our empirical study focuses on latent generative models based on diffusion and flow matching. While latent distribution mismatch may extend more broadly, we do not empirically investigate autoregressive latent generators or other generation frameworks.

Second, our explanation for the relationship between GAR-FID and generative performance is primarily empirical. Although we observe consistent associations among GAR-FID, intermediate latent mismatch, final latent mismatch, and gFID, we do not provide formal theoretical guarantees for these relationships. Proposition~\ref{prop:latent-fd-gfid-bound} provides a complementary perspective under Gaussian and Lipschitz assumptions.

Third, our latent distribution analysis relies on comparisons within a shared latent space. The relationship between Latent-FD and gFID is most directly examined in fixed-tokenizer settings, where latent distributions are comparable across generators. While broader experiments show that the GAR-FID--gFID relationship remains strong when both tokenizers and generators vary, direct distribution-level comparisons across tokenizer latent spaces remain challenging.

Fourth, decoder adaptation is explicitly generation-oriented and need not simultaneously improve reconstruction quality. Adapting the decoder toward generation-relevant latent distributions can trade reconstruction performance under $\mathcal{P}_e$ for improved behavior under generation-time latents.

Finally, GAR and decoder adaptation introduce additional computation by passing intermediate latents through the generative model during evaluation or training. GAR cost depends on the underlying generative process and noise level: for multi-step generators, intermediate GAR evaluations can require fewer function evaluations than full generation, although this advantage is not universal. Decoder adaptation also depends on the intermediate noise range, reflecting a trade-off between generation awareness and source correspondence that may require tuning across model families.

GAR-FID should therefore be viewed as a trajectory-based diagnostic rather than a replacement for direct generative evaluation. It reveals how decoder behavior evolves from reconstruction toward generation and correlates strongly with gFID, but gFID remains necessary when final sample quality is the primary evaluation target.

\section{Additional Results for Cross-Tokenizer and Cross-Model Generalization}
\label{appendix:additional_generalization_results}

To examine whether the relationship observed in the controlled setting (Section~\ref{sec:controlled_analysis}) remains stable when both the tokenizer and generative model vary, we follow iFID~\citep{Xu2026MakingRF} and consider 14 VAEs with publicly available checkpoints. Among them, 13 are evaluated with both SiT-B and SiT-XL, while RAE is evaluated with SiT-XL only. The details of these models are provided in Table~\ref{tab:vaes}. These VAEs cover a diverse set of design choices. Some models, such as SD-VAE~\citep{Rombach2022HighResolutionIS}, IN-VAE~\citep{Leng2025REPAEUV}, QW-VAE~\citep{wu2025qwen}, FLUX-VAE~\citep{flux2024}, and SD3-VAE~\citep{esser2024scaling}, are optimized primarily for reconstruction. Others incorporate additional inductive biases, such as equivariance regularization in EQ-VAE~\citep{Kouzelis2025EQVAEER}. A third group leverages contrastive learning-based image encoders, including VA-VAE~\citep{Yao2025ReconstructionVG}, SOFT-VQ~\citep{Chen2024SoftVQVAEE1}, MAE-TOK~\citep{Chen2025MaskedAA}, DE-TOK~\citep{Chen2025MaskedAA}, DM-VAE~\citep{Ye2025DistributionMV}, REPAE-VAE~\citep{Leng2025REPAEUV}, and RAE~\citep{zheng2026diffusion}.

For each VAE, we use the publicly available SiT-B and SiT-XL~\citep{Ma2024SiTEF} generative models released by iFID~\citep{Xu2026MakingRF}, which are trained on the $256 \times 256$ ImageNet dataset~\citep{Deng2009ImageNetAL} in the corresponding latent space. We directly use their provided checkpoints without additional retraining. For each tokenizer-generator configuration, we compute rFID, GAR-FID (without CFG) across noise levels $\eta_t \in [0.1, 1.0]$, and gFID both with and without classifier-free guidance (CFG). The results are summarized in Table~\ref{tab:sit-b} for SiT-B and Table~\ref{tab:sit-xl} for SiT-XL.

\paragraph{Note on FID references.}
As shown in Table~\ref{tab:sit-b} and Table~\ref{tab:sit-xl}, GAR-FID (without CFG) at $\eta_t = 1.0$ is not equal to gFID without CFG. This discrepancy arises because the two metrics are evaluated with different reference image statistics. Specifically, gFID is computed using the precomputed ImageNet statistics (\texttt{VIRTUAL\_imagenet256\_labeled.npz})~\citep{Deng2009ImageNetAL}, whereas GAR-FID is computed against the 50K input-image PNG reference set used in the paired reconstruction protocol. As a result, GAR-FID at $\eta_t = 1.0$ is not expected to exactly match gFID without CFG.

\begin{table}[!t]
\caption{\small{Quantitative results for 13 tokenizers. 
For each tokenizer, we evaluate rFID, gFID (w/o and w/ CFG), 
and GAR-FID without CFG at noise levels $\eta_t$ ranging from 0.1 to 1.0. 
All results are obtained using SiT/B models trained on the corresponding tokenizer~\citep{Xu2026MakingRF}.}}
\vspace{-2ex}
\label{tab:sit-b}
\resizebox{\linewidth}{!}{
\begin{tabular}{@{}lccccccccccccc@{}}
\toprule
\multirow{2}{*}{Tokenizers} & \multirow{2}{*}{rFID ({\color{green!60!black}\bfseries$\pmb{\downarrow}$})} & \multicolumn{2}{c}{gFID ({\color{green!60!black}\bfseries$\pmb{\downarrow}$})} & \multicolumn{10}{c}{GAR-FID ($\eta_t$) ({\color{green!60!black}\bfseries$\pmb{\downarrow}$})}\\ \cmidrule(lr){3-4} \cmidrule(lr){5-14} & & w/o CFG & w CFG & 0.1 & 0.2 & 0.3 & 0.4 & 0.5 & 0.6 & 0.7 & 0.8 & 0.9 & 1.0 \\\midrule
SD-VAE~\citep{Rombach2022HighResolutionIS} & 0.73 & 46.64 & 9.91 & 1.16 & 2.28 & 4.85 & 9.63 & 17.10 & 25.94 & 33.98 & 38.91 & 41.35 & 42.03 \\
IN-VAE~\citep{Leng2025REPAEUV} & 0.26 & 48.28 & 10.87 & 0.46 & 0.88 & 1.86 & 3.86 & 7.68 & 14.53 & 24.37 & 35.13 & 41.78 & 43.36 \\
FLUX-VAE~\citep{flux2024} & 0.16 & 62.92 & 14.38 & 1.05 & 2.32 & 5.42 & 11.09 & 18.97 & 29.14 & 39.71 & 49.25 & 55.71 & 58.33 \\
QwenImage-VAE~\citep{wu2025qwen} & 1.52 & 48.03 & 10.97 & 2.14 & 2.94 & 4.54 & 7.14 & 11.17 & 16.84 & 24.20 & 32.57 & 40.49 & 43.72 \\
SD3-VAE~\citep{esser2024scaling} & 0.21 & 51.58 & 11.89 & 0.59 & 1.06 & 2.25 & 4.76 & 9.34 & 16.36 & 26.03 & 35.92 & 43.70 & 46.82 \\
EQ-VAE~\citep{Kouzelis2025EQVAEER} & 0.59 & 37.49 & 9.32 & 3.37 & 6.54 & 10.11 & 13.91 & 17.79 & 21.83 & 26.12 & 29.70 & 32.06 & 33.26 \\
VA-VAE-16~\citep{Yao2025ReconstructionVG} & 0.30 & 18.88 & 6.01 & 2.52 & 3.71 & 4.91 & 6.21 & 7.46 & 8.85 & 10.48 & 12.57 & 14.90 & 15.78 \\
VA-VAE-64~\citep{Yao2025ReconstructionVG} & 0.15 & 33.56 & 7.76 & 1.90 & 3.10 & 5.02 & 7.44 & 10.32 & 13.48 & 17.07 & 21.10 & 25.98 & 29.50 \\
SOFT-VQ~\citep{Chen2024SoftVQVAEE1} & 0.59 & 29.48 & 7.48 & 1.64 & 5.63 & 10.52 & 15.19 & 18.97 & 21.93 & 24.13 & 25.38 & 25.93 & 26.07 \\
MAE-TOK~\citep{Chen2025MaskedAA} & 0.61 & 13.75 & 5.69 & 1.66 & 3.42 & 4.95 & 6.34 & 7.55 & 8.64 & 9.93 & 10.89 & 11.56 & 11.64 \\
DE-TOK~\citep{yang2026latent} & 0.59 & 20.17 & 6.95 & 0.63 & 0.72 & 0.91 & 1.30 & 2.03 & 3.48 & 6.74 & 12.05 & 16.16 & 16.87 \\
DM-VAE~\citep{Ye2025DistributionMV} & 0.72 & 8.38 & 4.89 & 0.81 & 0.87 & 0.95 & 1.05 & 1.20 & 1.47 & 2.10 & 3.60 & 5.75 & 6.76 \\
REPAE-VAE~\citep{Leng2025REPAEUV} & 0.56 & 25.75 & 6.46 & 0.94 & 1.78 & 3.10 & 4.91 & 7.08 & 9.86 & 13.12 & 17.18 & 21.04 & 22.14 \\
\bottomrule
\end{tabular}
}
\vspace{-2ex}
\end{table}

\begin{table}[!t]
\caption{\small{Quantitative results for 13 tokenizers. 
For each tokenizer, we evaluate rFID, gFID (w/o and w/ CFG), 
and GAR-FID without CFG at noise levels $\eta_t$ ranging from 0.1 to 1.0. 
All results are obtained using SiT/XL models trained on the corresponding tokenizer~\citep{Xu2026MakingRF}.}}
\vspace{-2ex}
\label{tab:sit-xl}
\resizebox{\linewidth}{!}{
\begin{tabular}{@{}lccccccccccccc@{}}
\toprule
\multirow{2}{*}{Tokenizers} & \multirow{2}{*}{rFID ({\color{green!60!black}\bfseries$\pmb{\downarrow}$})} & \multicolumn{2}{c}{gFID ({\color{green!60!black}\bfseries$\pmb{\downarrow}$})} & \multicolumn{10}{c}{GAR-FID ($\eta_t$) ({\color{green!60!black}\bfseries$\pmb{\downarrow}$})}\\ \cmidrule(lr){3-4} \cmidrule(lr){5-14} & & w/o CFG & w CFG & 0.1 & 0.2 & 0.3 & 0.4 & 0.5 & 0.6 & 0.7 & 0.8 & 0.9 & 1.0 \\\midrule
SD-VAE~\citep{Rombach2022HighResolutionIS} & 0.73 & 25.91 & 6.33 & 1.11 & 1.76 & 2.91 & 5.03 & 8.27 & 12.82 & 17.31 & 20.43 & 21.85 & 22.30 \\
IN-VAE~\citep{Leng2025REPAEUV} & 0.26 & 25.65 & 6.56 & 0.45 & 0.69 & 1.15 & 1.97 & 3.48 & 6.18 & 10.75 & 16.65 & 20.58 & 21.51 \\
FLUX-VAE~\citep{flux2024} & 0.16 & 34.06 & 7.82 & 0.96 & 1.66 & 3.13 & 5.54 & 8.87 & 13.45 & 18.62 & 24.00 & 27.85 & 29.65 \\
QwenImage-VAE~\citep{wu2025qwen} & 1.52 & 23.62 & 6.19 & 2.13 & 2.61 & 3.33 & 4.35 & 5.82 & 7.88 & 10.76 & 14.52 & 18.29 & 19.92 \\
SD3-VAE~\citep{esser2024scaling} & 0.21 & 26.38 & 6.31 & 0.59 & 0.90 & 1.51 & 2.59 & 4.50 & 7.38 & 11.62 & 16.91 & 21.27 & 22.89 \\
EQ-VAE~\citep{Kouzelis2025EQVAEER} & 0.59 & 20.81 & 6.24 & 2.53 & 4.14 & 5.79 & 7.39 & 9.29 & 11.32 & 13.37 & 15.59 & 16.99 & 17.53 \\
VA-VAE-16~\citep{Yao2025ReconstructionVG} & 0.30 & 8.57 & 4.20 & 2.11 & 2.75 & 3.30 & 3.85 & 4.32 & 4.88 & 5.49 & 6.30 & 7.32 & 7.80 \\
VA-VAE-64~\citep{Yao2025ReconstructionVG} & 0.15 & 15.40 & 5.36 & 1.67 & 2.35 & 3.23 & 4.28 & 5.51 & 6.82 & 8.25 & 10.03 & 12.05 & 13.33 \\
SOFT-VQ~\citep{Chen2024SoftVQVAEE1} & 0.59 & 15.88 & 5.14 & 1.29 & 3.39 & 5.83 & 8.01 & 9.92 & 11.56 & 12.65 & 13.37 & 13.62 & 13.62 \\
MAE-TOK~\citep{Chen2025MaskedAA} & 0.61 & 6.27 & 3.74 & 1.23 & 1.99 & 2.59 & 3.08 & 3.54 & 3.94 & 4.44 & 4.91 & 5.19 & 5.21 \\
DE-TOK~\citep{yang2026latent} & 0.59 & 11.97 & 4.83 & 0.65 & 0.73 & 0.88 & 1.13 & 1.52 & 2.26 & 3.77 & 6.58 & 9.14 & 9.59 \\
DM-VAE~\citep{Ye2025DistributionMV} & 0.72 & 4.65 & 3.39 & 0.81 & 0.86 & 0.92 & 1.00 & 1.09 & 1.24 & 1.50 & 2.12 & 3.17 & 3.86 \\
REPAE-VAE~\citep{Leng2025REPAEUV} & 0.56 & 12.95 & 4.17 & 0.90 & 1.36 & 1.94 & 2.72 & 3.57 & 4.64 & 6.00 & 7.85 & 9.89 & 10.60 \\
RAE~\citep{zheng2026diffusion} & 0.63 & 4.25 & 3.50 & 0.64 & 0.68 & 0.82 & 0.89 & 0.99 & 1.10 & 1.25 & 1.47 & 1.92 & 4.38 \\
\bottomrule
\end{tabular}
}
\vspace{-1ex}
\end{table}

\begin{table}[!t]
\caption{\small{List of VAEs included in our study. We follow iFID~\citep{Xu2026MakingRF} and use publicly available generative-model checkpoints trained in their corresponding latent spaces.}}
\vspace{-2ex}
\label{tab:vaes}
\centering
\resizebox{0.75\linewidth}{!}{
\begin{tabular}{@{}lccc@{}}
\toprule
VAE Name & Latent Dim. & Arch. & Training  \\ \midrule
SD-VAE \citep{Rombach2022HighResolutionIS} & $4\times 32 \times 32$ & UNet & Recon.  \\
IN-VAE \citep{Leng2025REPAEUV} & $16\times 16 \times 16$ & UNet & Recon.  \\
FLUX-VAE \citep{flux2024} & $16\times 32\times 32$ & UNet & Recon. \\
QwenImage-VAE \citep{wu2025qwen} & $16\times 32 \times 32$ & UNet & Recon. \\
SD3-VAE \citep{esser2024scaling} & $16\times 32 \times 32$ & UNet & Recon.  \\
EQ-VAE \citep{Kouzelis2025EQVAEER} & $4\times 32 \times 32$ & UNet & Recon. + Equivariance  \\
VA-VAE-16 \citep{Yao2025ReconstructionVG} & $16\times 16 \times 16$ & UNet & Recon. + DINO alignment  \\
VA-VAE-64 \citep{Yao2025ReconstructionVG} & $64\times 16 \times 16$ & UNet & Recon. + DINO alignment  \\
SOFT-VQ \citep{Chen2024SoftVQVAEE1} & $64\times 32$ & ViT & Recon. + DINO alignment  \\
MAE-TOK \citep{Chen2025MaskedAA} & $128\times 32$ & ViT & Recon. + Mask + DINO alignment  \\
DE-TOK \citep{yang2026latent} & $128\times 32$ & ViT & Recon. + Mask + latent denoising  \\
DM-VAE \citep{Ye2025DistributionMV} & $256\times 32$ & ViT & Recon. + Distribution Matching  \\
REPAE-VAE \citep{Leng2025REPAEUV} & $4\times 32 \times 32$ & UNet & Recon. + REPA Loss  \\
RAE \citep{zheng2026diffusion} & $768\times 16 \times 16$ & ViT & DINO Encoder + Recon. Decoder  \\
\bottomrule
\end{tabular}
}
\end{table}

\begin{table}[!t]
\caption{\small{Correlation between FID-based metrics and generative performance (gFID). 
Correlations are computed across models obtained by training SiT with 13 pre-trained VAEs. We report both Pearson correlation (PCC) and Spearman rank correlation (SRCC) under different settings (with and without CFG) for SiT-B, SiT-XL, and their mixed configurations. GAR-FID without CFG consistently exhibits stronger correlation with gFID across noise levels $\eta_t$, whereas the standard reconstruction metric (rFID) shows negative correlation. \textbf{Bold} indicates the best results. $^{\dagger}$ denotes results from~\citep{Xu2026MakingRF}, and $^{\star}$ indicates our reproduced results.  
}}
\vspace{-2ex}
\label{tab:resultfid-full}
\resizebox{\linewidth}{!}{
\begin{tabular}{@{}lcccccccccccc@{}}
\toprule
\multirow{3}{*}{Metrics} & \multicolumn{4}{c}{gFID SiT/B} & \multicolumn{4}{c}{gFID SiT/XL} & \multicolumn{4}{c}{gFID SiT/mixed}\\
\cmidrule(lr){2-5}\cmidrule(lr){6-9}\cmidrule(lr){10-13}
 & \multicolumn{2}{c}{w/o CFG} & \multicolumn{2}{c}{w/ CFG} & \multicolumn{2}{c}{w/o CFG} & \multicolumn{2}{c}{w/ CFG} & \multicolumn{2}{c}{w/o CFG} & \multicolumn{2}{c}{w/ CFG} \\
\cmidrule(lr){2-3}\cmidrule(lr){4-5}\cmidrule(lr){6-7}\cmidrule(lr){8-9}\cmidrule(lr){10-11}\cmidrule(lr){12-13}
 & PCC ({\color{green!60!black}\bfseries$\pmb{\uparrow}$}) & SRCC ({\color{green!60!black}\bfseries$\pmb{\uparrow}$}) & PCC ({\color{green!60!black}\bfseries$\pmb{\uparrow}$}) & SRCC ({\color{green!60!black}\bfseries$\pmb{\uparrow}$}) & PCC ({\color{green!60!black}\bfseries$\pmb{\uparrow}$}) & SRCC ({\color{green!60!black}\bfseries$\pmb{\uparrow}$}) & PCC ({\color{green!60!black}\bfseries$\pmb{\uparrow}$}) & SRCC ({\color{green!60!black}\bfseries$\pmb{\uparrow}$}) & PCC ({\color{green!60!black}\bfseries$\pmb{\uparrow}$}) & SRCC ({\color{green!60!black}\bfseries$\pmb{\uparrow}$}) & PCC ({\color{green!60!black}\bfseries$\pmb{\uparrow}$})& SRCC ({\color{green!60!black}\bfseries$\pmb{\uparrow}$}) \\
\midrule
rFID$^{\dagger}$ & -0.04 & -0.31 & -0.07 & -0.31 & -0.06 & -0.21 & -0.15 & -0.31 & -- & -- & -- & -- \\
iFID$^{\dagger}$ & 0.85 & 0.86 & 0.82 & 0.84 & 0.89 & 0.91 & 0.88 & 0.92 & -- & -- & -- & -- \\
rFID$^{\star}$ & -0.11 & -0.36 & -0.11 & -0.30 & -0.13 & -0.34 & -0.15 & -0.39 & -0.10 & -0.30 & -0.10 & -0.28 \\
iFID$^{\star}$ & 0.86 & 0.83 & 0.80 & 0.79 & 0.90 & 0.88 & 0.88 & 0.90 & 0.72 & 0.77 & 0.64 & 0.70 \\
GAR-FID ($\eta_t=0.1$) & -0.10 & -0.19 & -0.13 & -0.14 & 0.01 & 0.00 & 0.10 & 0.01 & 0.04 & 0.03 & 0.05 & 0.01 \\
GAR-FID ($\eta_t=0.2$) & -0.04 & -0.04 & -0.08 & -0.04 & 0.12 & 0.22 & 0.19 & 0.20 & 0.16 & 0.25 & 0.16 & 0.22 \\
GAR-FID ($\eta_t=0.3$) & 0.14 & 0.15 & 0.09 & 0.14 & 0.25 & 0.38 & 0.32 & 0.35 & 0.34 & 0.42 & 0.34 & 0.39 \\
GAR-FID ($\eta_t=0.4$) & 0.38 & 0.33 & 0.33 & 0.34 & 0.45 & 0.54 & 0.49 & 0.51 & 0.55 & 0.64 & 0.54 & 0.60 \\
GAR-FID ($\eta_t=0.5$) & 0.63 & 0.68 & 0.57 & 0.69 & 0.61 & 0.66 & 0.65 & 0.61 & 0.74 & 0.80 & 0.72 & 0.76 \\
GAR-FID ($\eta_t=0.6$) & 0.81 & 0.79 & 0.77 & 0.78 & 0.79 & 0.84 & 0.80 & 0.79 & 0.86 & 0.89 & 0.84 & 0.85 \\
GAR-FID ($\eta_t=0.7$) & 0.93 & 0.90 & 0.89 & 0.87 & 0.91 & 0.90 & 0.91 & 0.86 & 0.95 & 0.96 & 0.93 & 0.91 \\
GAR-FID ($\eta_t=0.8$) & 0.98 & 0.96 & 0.96 & 0.94 & 0.98 & 0.99 & 0.97 & 0.96 & 0.99 & 0.99 & 0.97 & 0.95 \\
GAR-FID ($\eta_t=0.9$) & \textbf{1.00} & \textbf{0.99} & \textbf{0.98} & 0.98 & \textbf{1.00} & \textbf{1.00} & \textbf{0.98} & \textbf{0.96} & \textbf{1.00} & \textbf{1.00} & \textbf{0.98} & \textbf{0.97} \\
GAR-FID ($\eta_t=1.0$) & \textbf{1.00} & \textbf{0.99} & \textbf{0.98} & \textbf{0.99} & \textbf{1.00} & \textbf{1.00} & \textbf{0.98} & \textbf{0.96} & \textbf{1.00} & \textbf{1.00} & \textbf{0.98} & \textbf{0.97} \\
\bottomrule
\end{tabular}
}
\vspace{-4ex}
\end{table}
To quantify the relationship between these metrics and generative performance, we follow iFID~\citep{Xu2026MakingRF} and compute the Pearson correlation coefficient (PCC) and Spearman rank correlation coefficient (SRCC) between each metric and gFID, both with and without CFG. As reported in Table~\ref{tab:resultfid-full}, rFID exhibits consistently negative correlation with gFID across SiT-B, SiT-XL, and their mixed configurations. This indicates that better reconstruction FID can correspond to worse generative FID, consistent with the reconstruction-generation dilemma observed in prior work~\citep{Yao2025ReconstructionVG,Kouzelis2025EQVAEER,Ye2025DistributionMV,Skorokhodov2025ImprovingTD,Chen2025MaskedAA,Xu2026MakingRF}. In contrast, the correlation between GAR-FID and gFID strengthens substantially as the noise level $\eta_t$ increases. At $\eta_t=0.8$, GAR-FID already achieves consistently high correlation across settings, with SRCC ranging from $0.94$ to $0.99$. When $\eta_t\geq0.8$, GAR-FID also outperforms iFID~\citep{Xu2026MakingRF} on both SiT-B and SiT-XL across the reported correlation settings, with the correlation approaching saturation at higher noise levels.

The mixed SiT-B/XL setting further tests whether the GAR-FID-gFID relationship remains stable across generator scales. Correlations computed within a single generative model, such as SiT-B or SiT-XL~\citep{Ma2024SiTEF}, primarily capture tokenizer variation under a fixed generator scale. By contrast, the mixed setting jointly varies tokenizer choice and generator scale, providing a more demanding test of cross-system comparisons. In this setting, the correlation of iFID with gFID decreases, whereas GAR-FID maintains consistently strong correlation at higher noise levels. These results indicate that the GAR-FID--gFID relationship remains robust across heterogeneous tokenizer-generator configurations.

\paragraph{Leave-one-out robustness across tokenizers.}
To test whether the strong correlation between GAR-FID and gFID is driven by any individual tokenizer, we perform a leave-one-out analysis using all 14 tokenizer configurations in Table~\ref{tab:sit-xl}. Specifically, we omit each tokenizer in turn and recompute the SRCC over the remaining 13 configurations. Tables~\ref{tab:loo-srcc-nocfg} and~\ref{tab:loo-srcc-cfg} report the complete SRCC results without and with CFG, respectively.

At low noise levels, particularly when $\eta_t<0.5$, the correlations are relatively weak and more sensitive to the removal of individual tokenizers. They increase substantially as $\eta_t$ grows and become uniformly high at $\eta_t\geq0.8$. In particular, at $\eta_t=0.8$, the leave-one-out SRCC is $0.99$ for every omitted tokenizer without CFG and ranges from $0.96$ to $0.97$ with CFG. This demonstrates that the strong GAR-FID--gFID relationship at $\eta_t=0.8$ is not driven by any single tokenizer. 

\begin{table*}[!t]
\centering
\caption{\small{Leave-one-out SRCC between GAR-FID and gFID without CFG. Each column
omits the indicated tokenizer when computing the correlation.}}
\vspace{-2ex}
\label{tab:loo-srcc-nocfg}
\scriptsize
\resizebox{\linewidth}{!}{
\begin{tabular}{@{}lcccccccccccccc@{}}
\toprule
Omitted tokenizer & SD-VAE & IN-VAE & FLUX-VAE & QwenImage-VAE & SD3-VAE & EQ-VAE & VA-VAE-16 & VA-VAE-64 & SOFT-VQ & MAE-TOK & DE-TOK & DM-VAE & REPAE-VAE & RAE \\
\midrule
GAR-FID ($\eta_t=0.1$) & 0.03 & 0.18 & 0.00 & -0.03 & 0.21 & -0.03 & 0.05 & -0.01 & -0.01 & 0.02 & -0.05 & -0.11 & -0.07 & -0.13 \\
GAR-FID ($\eta_t=0.2$) & 0.22 & 0.31 & 0.21 & 0.20 & 0.25 & 0.16 & 0.26 & 0.17 & 0.16 & 0.21 & 0.10 & 0.03 & 0.12 & -0.04 \\
GAR-FID ($\eta_t=0.3$) & 0.38 & 0.45 & 0.37 & 0.37 & 0.47 & 0.37 & 0.44 & 0.37 & 0.37 & 0.32 & 0.27 & 0.21 & 0.32 & 0.19 \\
GAR-FID ($\eta_t=0.4$) & 0.52 & 0.65 & 0.48 & 0.55 & 0.68 & 0.58 & 0.55 & 0.51 & 0.58 & 0.55 & 0.47 & 0.42 & 0.54 & 0.42 \\
GAR-FID ($\eta_t=0.5$) & 0.66 & 0.79 & 0.64 & 0.69 & 0.70 & 0.73 & 0.65 & 0.65 & 0.73 & 0.64 & 0.62 & 0.58 & 0.63 & 0.58 \\
GAR-FID ($\eta_t=0.6$) & 0.85 & 0.90 & 0.82 & 0.89 & 0.90 & 0.89 & 0.85 & 0.85 & 0.89 & 0.84 & 0.85 & 0.82 & 0.84 & 0.82 \\
GAR-FID ($\eta_t=0.7$) & 0.90 & 0.92 & 0.88 & 0.92 & 0.93 & 0.93 & 0.89 & 0.88 & 0.93 & 0.89 & 0.90 & 0.88 & 0.88 & 0.88 \\
GAR-FID ($\eta_t=0.8$) & 0.99 & 0.99 & 0.99 & 0.99 & 0.99 & 0.99 & 0.99 & 0.99 & 0.99 & 0.99 & 0.99 & 0.99 & 0.99 & 0.99 \\
GAR-FID ($\eta_t=0.9$) & 1.00 & 0.99 & 0.99 & 0.99 & 1.00 & 0.99 & 0.99 & 0.99 & 0.99 & 0.99 & 0.99 & 0.99 & 0.99 & 0.99 \\
GAR-FID ($\eta_t=1.0$) & 0.99 & 0.99 & 0.99 & 0.99 & 0.99 & 0.99 & 0.99 & 0.99 & 0.99 & 0.99 & 0.99 & 1.00 & 0.99 & 1.00 \\
\bottomrule
\end{tabular}
}
\vspace{-1ex}
\end{table*}

\begin{table*}[!t]
\centering
\caption{\small{Leave-one-out SRCC between GAR-FID and gFID with CFG. Each column
omits the indicated tokenizer when computing the correlation.}}
\vspace{-2ex}
\label{tab:loo-srcc-cfg}
\scriptsize
\resizebox{\linewidth}{!}{
\begin{tabular}{@{}lcccccccccccccc@{}}
\toprule
Omitted tokenizer & SD-VAE & IN-VAE & FLUX-VAE & QwenImage-VAE & SD3-VAE & EQ-VAE & VA-VAE-16 & VA-VAE-64 & SOFT-VQ & MAE-TOK & DE-TOK & DM-VAE & REPAE-VAE & RAE \\
\midrule
GAR-FID ($\eta_t=0.1$) & 0.04 & 0.23 & 0.01 & -0.03 & 0.20 & -0.03 & 0.03 & -0.01 & -0.01 & 0.02 & -0.02 & -0.08 & -0.08 & -0.10 \\
GAR-FID ($\eta_t=0.2$) & 0.20 & 0.34 & 0.17 & 0.15 & 0.23 & 0.12 & 0.19 & 0.15 & 0.14 & 0.17 & 0.10 & 0.03 & 0.07 & -0.05 \\
GAR-FID ($\eta_t=0.3$) & 0.35 & 0.45 & 0.33 & 0.34 & 0.43 & 0.34 & 0.38 & 0.36 & 0.36 & 0.29 & 0.27 & 0.19 & 0.29 & 0.17 \\
GAR-FID ($\eta_t=0.4$) & 0.49 & 0.66 & 0.45 & 0.51 & 0.63 & 0.54 & 0.51 & 0.51 & 0.58 & 0.51 & 0.46 & 0.40 & 0.51 & 0.40 \\
GAR-FID ($\eta_t=0.5$) & 0.61 & 0.79 & 0.57 & 0.61 & 0.64 & 0.64 & 0.58 & 0.61 & 0.68 & 0.58 & 0.56 & 0.52 & 0.58 & 0.52 \\
GAR-FID ($\eta_t=0.6$) & 0.81 & 0.89 & 0.77 & 0.84 & 0.85 & 0.84 & 0.79 & 0.83 & 0.86 & 0.79 & 0.81 & 0.78 & 0.79 & 0.78 \\
GAR-FID ($\eta_t=0.7$) & 0.85 & 0.91 & 0.82 & 0.87 & 0.88 & 0.86 & 0.85 & 0.85 & 0.90 & 0.84 & 0.86 & 0.83 & 0.85 & 0.83 \\
GAR-FID ($\eta_t=0.8$) & 0.97 & 0.97 & 0.96 & 0.96 & 0.97 & 0.96 & 0.97 & 0.96 & 0.96 & 0.96 & 0.97 & 0.96 & 0.97 & 0.96 \\
GAR-FID ($\eta_t=0.9$) & 0.96 & 0.97 & 0.95 & 0.96 & 0.96 & 0.96 & 0.96 & 0.96 & 0.96 & 0.95 & 0.96 & 0.96 & 0.97 & 0.96 \\
GAR-FID ($\eta_t=1.0$) & 0.97 & 0.97 & 0.95 & 0.96 & 0.97 & 0.96 & 0.96 & 0.96 & 0.96 & 0.95 & 0.96 & 0.95 & 0.97 & 0.95 \\
\bottomrule
\end{tabular}
}
\vspace{-2ex}
\end{table*}

\section{Decoder Feature Spaces for Latent Distribution Analysis}
\label{appendix:decoder space}
In Section~\ref{sec:latent mismatch}, we analyze intermediate representations from the SD-VAE decoder~\citep{Rombach2022HighResolutionIS}. Specifically, we extract features from three representative decoder stages:
(i) \emph{Conv In}, the initial convolutional projection of the latent input;
(ii) \emph{Mid Block}, the bottleneck stage at the lowest spatial resolution;
and (iii) \emph{Up Block 0}, the first upsampling stage of the decoder.

These feature spaces correspond to progressively deeper stages of the decoding process and are identified from the official implementation. We collect activations using forward hooks at \texttt{decoder.conv\_in}, \texttt{decoder.mid\_block}, and \texttt{decoder.up\_blocks[0]} for both encoder and generator latents. For visualization, activations from each feature space are adaptively pooled to a fixed $4\times4$ spatial resolution, flattened, and projected onto a one-dimensional LDA direction fitted to distinguish encoder-induced from generator-induced representations.

\section{gFID Evaluation Protocol and Implementation Details}
\label{appendix:gfid_protocol}

\begin{table}[!h]
\centering
\vspace{-2ex}
\caption{\small{gFID is sensitive to the evaluation protocol.
We compare two protocols on iMF-B/2: the official iMF evaluation~\citep{Geng2025ImprovedMF} and the OpenAI protocol.
They differ in computational platform (TPU vs.\ GPU), Inception network implementation and weights (PyTorch vs.\ TensorFlow), and reference image statistics.}}
\vspace{-2ex}
\label{tab:fid_protocol_sensitivity}
\small
\resizebox{\linewidth}{!}{
\begin{tabular}{@{}lccccc@{}}
\toprule
Setting & Platform & Inception & Reference Image Statistics & FID & IS \\
\midrule
Official iMF & TPU & PyTorch & \texttt{jit\_in256\_stats.npz} & 3.39 & 255.3 \\
Official iMF (our reproduction) & GPU & PyTorch & \texttt{jit\_in256\_stats.npz} & 3.37 & 255.7 \\
OpenAI Setup & GPU & TensorFlow & \texttt{VIRTUAL\_imagenet256\_labeled.npz} & 3.47 & 255.2 \\
\bottomrule
\end{tabular}
}
\vspace{-2ex}
\end{table}

The official iMF-B/2~\citep{Geng2025ImprovedMF} results are obtained under a different setup from the commonly used OpenAI evaluation protocol.
Specifically, the two protocols differ in three aspects:
(i) computational platform (TPU vs.\ GPU),
(ii) Inception network implementation and weights (PyTorch vs.\ TensorFlow),
and (iii) reference image statistics.

Table~\ref{tab:fid_protocol_sensitivity} summarizes these differences and their impact on reported FID scores. We observe that replacing the TPU platform with GPU, while keeping the other components unchanged, yields slightly improved FID in our reproduction. However, further adopting the OpenAI evaluation setup by switching to the TensorFlow Inception implementation and using different reference image statistics leads to noticeably worse FID. This indicates that the commonly used OpenAI evaluation protocol tends to produce systematically higher (worse) FID scores. This observation is also consistent with FD-loss~\citep{Yang2026RepresentationFL}, who similarly report higher FID scores under the OpenAI evaluation protocol.

Unless otherwise specified, we adopt the commonly used OpenAI evaluation protocol (TensorFlow Inception network with \texttt{VIRTUAL\_imagenet256\_labeled.npz} statistics) for all reported gFID results, to ensure consistency and comparability across different models and prior works.

\section{Experimental Details in Section~\ref{sec:experiments}}
\label{app:exp_details}

\paragraph{Data Preprocessing.}
All experiments are conducted on ImageNet-1k~\citep{Deng2009ImageNetAL}. 
Following LlamaGen\\~\citep{Sun2024AutoregressiveMB}, we apply iterative box downsampling to resize all images to a resolution of $256 \times 256$.

\paragraph{Training Details.}
We perform decoder adaptation on four latent generative models, iMF-B/2, iMF-M/2, iMF-L/2, and iMF-XL/2~\citep{Geng2025ImprovedMF}, which share a common tokenizer, SD-VAE~\citep{Rombach2022HighResolutionIS}. During adaptation, the tokenizer encoder and latent generative model are kept frozen, and only the tokenizer decoder parameters are updated. We consider two adaptation settings. In the standard setting, intermediate GAR latents are generated without classifier-free guidance (CFG), and the decoder is adapted for 10 epochs. We additionally evaluate a \emph{CFG-aware decoder adaptation} setting, in which intermediate GAR latents are generated with CFG to better match the guided generation regime used at inference time. CFG-aware adaptation is performed for one epoch, while all other training settings remain unchanged. To further evaluate the generality of decoder adaptation beyond iMF, we apply the same one-epoch CFG-aware adaptation setting to MeanFlow-XL/2~\citep{Geng2025MeanFF} and the distillation-based DMF-XL/2+~\citep{Lee2025DecoupledMT}. All experiments are conducted on 8 NVIDIA H100 GPUs (80GB memory each) using the AdamW optimizer~\citep{Loshchilov2017DecoupledWD} with $\beta_1=0.9$ and $\beta_2=0.95$. We use a fixed learning rate of $10^{-5}$, a weight decay of $10^{-4}$, and a batch size of 32 per GPU. Under the standard 10-epoch setting, training typically completes within approximately 12 hours for iMF-B/2 and 14 hours for iMF-XL/2.

\paragraph{Adversarial Training.} For adversarial supervision used in Section~\ref{sec:method}, we follow established visual-tokenizer training practices~\citep{Tian2024VisualAM,Chen2024SoftVQVAEE1,Fang2025VQTransplant,Li2024ImageFolderAI} and employ a discriminator built on a frozen DINO-S backbone~\citep{Caron2021EmergingPI,Oquab2023DINOv2LR}. We further use differentiable augmentation (DiffAug)~\citep{Zhao2020DifferentiableAF}, consistency regularization~\citep{Zhang2019ConsistencyRF}, and LeCAM regularization~\citep{Tseng2021RegularizingGA} for adversarial training. These training choices and loss weights are kept fixed across model scales.

\paragraph{Loss Weights and Noise Level.}
We fix $\lambda_1 = 1.0$, $\lambda_2 = 1.0$, $\lambda_3 = 0.7$, and $\lambda_4 = 10.0$ across all experiments. 
For noise level selection in decoder adaptation, we sample $\eta_t \sim \mathrm{Uniform}[\eta_{\min}, \eta_{\max}]$, where $\eta_{\min} = 0.25$ and $\eta_{\max} = 0.45$, as discussed in Appendix~\ref{appendix:noise_level_choice}.

\paragraph{CFG parameters.}  The official iMF implementation employs three classifier-free guidance (CFG) parameters: $t_{\min}$, $t_{\max}$, and $\omega$. In the original iMF repository, these parameters are selected via a fine-grained grid search. Since decoder adaptation is performed without CFG during training, the original CFG hyperparameters are no longer optimal after updating the decoder. We therefore conduct a new grid search to identify a new set of CFG parameters for evaluation. The updated CFG parameters are reported in Table~\ref{tab:CFG parameters}. We will release the corresponding checkpoints to further facilitate reproducibility by the community.

\begin{table}[!h]
\vspace{-2mm}
\caption{\small{CFG parameters used for gFID evaluation across different model scales. Left: official parameters from the iMF repository, obtained via fine-grained grid search without decoder adaptation. Right: re-optimized parameters after decoder adaptation, obtained via a new grid search to account for decoder changes.}}
\vspace{-2ex}
\label{tab:CFG parameters}
\resizebox{\linewidth}{!}{
\begin{tabular}{@{}lcccccccc@{}}
\toprule
\multirow{2}{*}{CFG Parameters} & \multicolumn{4}{c}{w/o Decoder Adaptation} & \multicolumn{4}{c}{ w/ Decoder Adaptation}\\
\cmidrule(lr){2-5} \cmidrule(lr){6-9}
& iMF-B/2 & iMF-M/2 & iMF-L/2 & iMF-XL/2 & iMF-B/2 & iMF-M/2 & iMF-L/2 & iMF-XL/2 \\
\midrule
$t_{\min}$ & 0.40 & 0.40 & 0.40 & 0.42 & 0.50 & 0.55 & 0.44 & 0.48 \\
$t_{\max}$ & 0.65 & 0.60 & 0.60 & 0.62 & 0.65 & 0.625 & 0.60 & 0.65 \\
$\omega$   & 8.0  & 10.5 & 10.5 & 8.0  & 9.0  & 10.5  & 8.0  & 5.25 \\
\bottomrule
\end{tabular}
}
\vspace{-1ex}
\end{table}

\section{Additional Experimental Results in Section~\ref{sec:experiments}}
\label{appendix:additional decoder adaptation}
We provide additional experimental results in this appendix due to space constraints in the main paper. Specifically, we report robustness analyses across random seeds, examine how decoder adaptation affects the relationship between reconstruction and generation, study sensitivity to adaptation noise levels, evaluate reconstruction performance, provide system-level comparisons with a broad range of generative models, and present additional qualitative visual results.

\paragraph{Robustness across random seeds.} To assess the robustness of decoder adaptation to random-seed variation, we repeat the gFID evaluation with three random seeds under the OpenAI evaluation protocol. As shown in Table~\ref{tab:seed_variance}, the standard deviation is consistently small across all model scales, both with and without CFG. Moreover, DA improves the mean gFID in every evaluated setting. Without CFG, the improvements are substantial across all four model scales, while with CFG the gains are smaller but remain consistent relative to the observed run-to-run variation. For example, on iMF-XL/2 with CFG, DA improves gFID from $1.81\pm0.03$ to $1.73\pm0.03$. These results indicate that the improvements from DA are stable across random seeds rather than being driven by a favorable evaluation run.

\begin{table}[!t]
\centering
\caption{\small{Repeated-run evaluation of decoder adaptation across three random seeds. We report mean $\pm$ standard deviation of gFID under the OpenAI evaluation protocol for four iMF model scales, with and without CFG, before and after decoder adaptation (DA).}}
\vspace{-2ex}
\label{tab:seed_variance}
\small
\resizebox{\linewidth}{!}{
\begin{tabular}{@{}lcccccccc@{}}
\toprule
\multirow{2}{*}{Setting} & \multicolumn{4}{c}{w/o CFG} & \multicolumn{4}{c}{w/ CFG} \\
\cmidrule(lr){2-5} \cmidrule(lr){6-9}
 & iMF-B/2 & iMF-M/2 & iMF-L/2 & iMF-XL/2 & iMF-B/2 & iMF-M/2 & iMF-L/2 & iMF-XL/2 \\
\midrule
w/o DA & $16.36\pm0.12$ & $11.79\pm0.04$ & $9.28\pm0.04$ & $9.52\pm0.17$ & $3.49\pm0.01$ & $2.39\pm0.02$ & $1.91\pm0.01$ & $1.81\pm0.03$ \\
w/ DA & $12.30\pm0.07$ & $9.51\pm0.01$ & $7.33\pm0.05$ & $7.48\pm0.09$ & $3.14\pm0.01$ & $2.27\pm0.03$ & $1.81\pm0.02$ & $1.73\pm0.03$ \\
\bottomrule
\end{tabular}
}
\vspace{-1ex}
\end{table}

\begin{figure}[!t]
\centering
\small
\includegraphics[width=\textwidth]{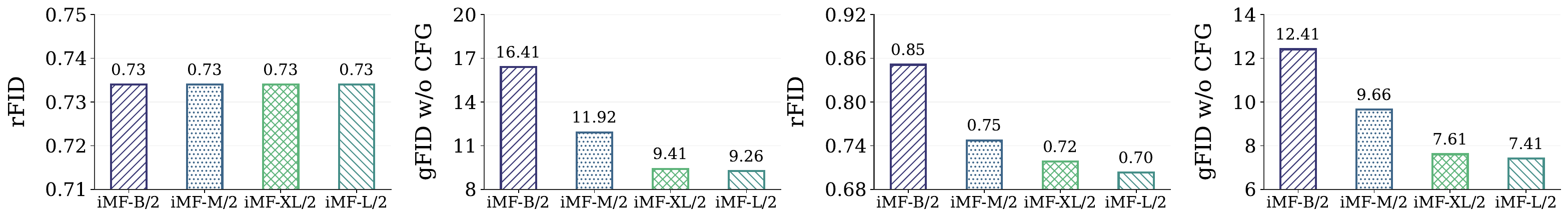}
\vspace{-7mm}
\caption{\small{rFID vs.\ gFID (w/o CFG) across model scales, before (left two) and after (right two) decoder adaptation. rFID is uninformative before adaptation but becomes aligned with gFID after adaptation.
}}
\label{fig:FID-post-train}
\vspace{-0.5mm}
\end{figure}

\begin{table}[!t]
\vspace{-2mm}
\centering
\caption{\small{\textbf{Effect of fixed noise levels in decoder adaptation.}
We report gFID without CFG, evaluated using the OpenAI protocol, and rFID across iMF model scales for decoder adaptation with different fixed noise levels $\eta_t$.
See Appendix~\ref{appendix:gfid_protocol} for protocol details.}}
\vspace{-1ex}
\label{tab:noise levels choices}
\resizebox{\linewidth}{!}{
\begin{tabular}{@{}lcccccccc@{}}
\toprule
\multirow{2}{*}{Noise Level} & \multicolumn{4}{c}{gFID w/o CFG ({\color{green!60!black}\bfseries$\pmb{\downarrow}$})} & \multicolumn{4}{c}{rFID  ({\color{green!60!black}\bfseries$\pmb{\downarrow}$})}\\
\cmidrule(lr){2-5} \cmidrule(lr){6-9}
& iMF-B/2 & iMF-M/2 & iMF-L/2 & iMF-XL/2 & iMF-B/2 & iMF-M/2 & iMF-L/2 & iMF-XL/2 \\
\midrule
$\eta_t =0.0$$^{\ddagger}$  & 16.41 & 11.92 & 9.26 & 9.41 & 0.73 & 0.73 & 0.73 & 0.73 \\
$\eta_t =0.0$  & 15.27 & 11.30 & 8.38 & 8.82 & \textbf{0.58} & 0.62 & 0.59 & 0.62 \\
$\eta_t =0.1$ & 14.76 & 11.03 & 8.19 & 8.64 & 0.62 & \textbf{0.61} & \textbf{0.58} & \textbf{0.60} \\
$\eta_t =0.2$ & 13.50 & 10.15 & 7.89 & 8.03 & 0.71 & 0.67 & 0.62 & 0.64 \\
$\eta_t =0.3$ & 12.86 & 9.99 & 7.57 & 7.84 & 0.77 & 0.70 & 0.67 & 0.67\\
$\eta_t =0.4$ & 12.08 & 9.52 & 7.51 & 7.24 & 0.94 & 0.79 & 0.75 & 0.74\\
$\eta_t =0.5$ & \textbf{11.58} & \textbf{9.18} & \textbf{7.36} & \textbf{7.19} & 1.20 & 1.01 & 0.94 & 0.89\\
$\eta_t =0.6$ & 11.88 & 9.34 & 7.41 & 7.40 & 1.81 & 1.42 & 1.26 & 1.21\\\midrule
\end{tabular}
}
\vspace{-2ex}
\end{table}

\paragraph{Effect of Decoder Adaptation on the rFID-gFID Relationship.} Figure~\ref{fig:FID-post-train} shows the relationship between rFID and gFID (w/o CFG) across model scales, before and after decoder adaptation. Before adaptation, rFID remains constant across models because the tokenizer is fixed, while gFID varies substantially with generator scale. After adaptation, rFID decreases together with gFID across model scales, indicating a closer correspondence between reconstruction and generation behavior. This trend is consistent with our latent distribution perspective: adapting the decoder on generation-relevant GAR latents changes its behavior under $\mathcal{P}_e$ in a generator-dependent manner, bringing reconstruction performance into closer agreement with downstream generative performance.

\paragraph{Sensitivity Analyses on Noise Levels.}
To study the effect of the noise level used for decoder adaptation, we conduct a sensitivity analysis by fixing $\eta_t \in \{0.0, 0.1, 0.2, 0.3, 0.4, 0.5, 0.6\}$ during adaptation. As shown in Table~\ref{tab:noise levels choices}, $\eta_t=0.5$ achieves the best gFID without CFG across all iMF model scales in this fixed-noise setting. When $\eta_t$ is too small, the intermediate GAR latents $z_g^t\sim\mathcal{P}_g^t$ remain relatively close to the encoder distribution $\mathcal{P}_e$, providing limited exposure to generation-relevant latent distributions and resulting in smaller gFID improvements. Increasing $\eta_t$ moves the decoder inputs further toward the generation-time regime, but also progressively weakens their correspondence with the source image $x$, making paired reconstruction supervision less reliable at large noise levels. Consistently, rFID increases monotonically with $\eta_t$ across all iMF model scales, reflecting the growing shift away from the reconstruction regime. These results highlight a trade-off between generation awareness and source correspondence and motivate using intermediate noise levels for decoder adaptation.

\paragraph{Reconstruction Results.}  As shown in Table~\ref{tab:post_train_reconstruction}, we compare reconstruction performance before and after decoder adaptation on the iMF-B/2 generative model, including peak signal-to-noise ratio (PSNR), structural similarity index (SSIM), Fréchet Inception Distance~\citep[FID;][]{Heusel2017GANsTB}, perceptual similarity~\citep[LPIPS;][]{Zhang2018TheUE}, and inception score~\citep[IS;][]{Salimans2016ImprovedTF}. All results are reported without CFG. Standard reconstruction (SR, $\eta_t=0$) degrades consistently after adaptation across all metrics, indicating that the adapted decoder is no longer well aligned with encoder latents $z_e$. 

For generation-aware reconstruction (GAR), we observe a systematic trade-off in the effect of decoder adaptation as the noise level $\eta_t$ increases. For structural fidelity, measured by PSNR and SSIM, decoder adaptation consistently degrades performance; however, the magnitude of degradation gradually diminishes with increasing $\eta_t$, indicating that the mismatch between the adapted decoder and the input latents becomes less severe as the latents move closer to the generative distribution. In contrast, perceptual quality, measured by LPIPS, IS, and FID, exhibits an opposite trend. At low noise levels, decoder adaptation slightly worsens performance, while at higher noise levels it yields consistent improvements. In particular, FID shows substantial gains when $\eta_t \geq 0.3$, suggesting that decoder adaptation improves perceptual alignment under generation-consistent latent distributions.

\paragraph{System-Level Comparison of Generative Performance.} We present a system-level comparison on class-conditional ImageNet $256\times256$ across diffusion/flow, GAN, and autoregressive or masked generative models, with inference cost measured by the number of function evaluations (NFE). For diffusion and flow models trained from scratch, we include 1-NFE methods MeanFlow-XL/2~\citep{Geng2025MeanFF}, TiM-XL/2~\citep{Wang2025TransitionMR}, $\alpha$-Flow-XL/2+~\citep{Zhang2025AlphaFlowUA}, and iMF~\citep{Geng2025ImprovedMF} at multiple scales (B/2, M/2, L/2, XL/2), together with 2-NFE methods IMM-XL/2~\citep{Zhou2025InductiveMM}, MeanFlow-XL/2+~\citep{Geng2025MeanFF}, $\alpha$-Flow-XL/2+~\citep{Zhang2025AlphaFlowUA}, and iMF-XL/2. We further include distillation-based 1-NFE methods $\pi$-Flow-XL/2~\citep{Chen2025piFlowPF}, DMF-XL/2+~\citep{Lee2025DecoupledMT}, and FACM-XL/2~\citep{Peng2025FlowAnchoredCM}, as well as multi-NFE diffusion/flow models ADM-G~\citep{Dhariwal2021DiffusionMB}, LDM-4-G~\citep{Rombach2022HighResolutionIS}, SimDiff~\citep{Hoogeboom2023simpleDE}, DiT-XL/2~\citep{Peebles2022ScalableDM}, SiT-XL/2~\citep{Ma2024SiTEF}, SiT-XL/2+REPA~\citep{Yu2024RepresentationAF}, SiD2~\citep{Hoogeboom2024SimplerD}, LightningDiT-XL/2~\citep{Yao2025ReconstructionVG}, DDT-XL/2~\citep{Wang2025DDTDD}, and RAE+DiT$^{\text{DH}}$-XL~\citep{zheng2026diffusion}. For broader reference, we also compare with GAN-based methods BigGAN~\citep{Brock2018LargeSG}, GigaGAN~\citep{Kang2023ScalingUG}, and StyleGAN-XL~\citep{Sauer2022StyleGANXLSS}, and autoregressive or masked models JetFormer-L~\citep{Tschannen2024JetFormerAA}, MaskGIT~\citep{Chang2022MaskGITMG}, RQ-Transformer~\citep{Lee2022AutoregressiveIG}, STARFlow~\citep{Gu2025STARFlowSL}, LLamaGen-3B~\citep{Sun2024AutoregressiveMB}, VAR-$d30$~\citep{Tian2024VisualAM}, MAR-H~\citep{Li2024AutoregressiveIG}, RAR-XXL~\citep{Yu2024RandomizedAV}, and xAR-H~\citep{Ren2025BeyondNN}.

As shown in Table~\ref{tab:all_results}, under the most constrained $\mathrm{NFE}=1$ setting, decoder adaptation consistently improves generative performance across all evaluated iMF scales. Importantly, the gains extend beyond iMF: applying the same decoder adaptation procedure to MeanFlow-XL/2 and the distillation-based DMF-XL/2+ also yields consistent improvements, indicating that the benefit is not specific to a particular generator architecture or training paradigm. Among the evaluated 1-NFE models, iMF-XL/2 + DA achieves the lowest gFID, while the adapted MeanFlow and DMF models also improve over their corresponding baselines. These results highlight decoder adaptation as a lightweight yet effective mechanism for improving generative quality without introducing additional inference cost.

\begin{table}[!t]
\caption{Reconstruction performance on the iMF-B/2 generative model. We compare reconstruction results without and with decoder adaptation. SR denotes standard reconstruction ($\eta_t = 0$), while GAR corresponds to generation-aware reconstruction with noise levels $\eta_t \in [0.1, 0.5]$.}
\vspace{-2ex}
\label{tab:post_train_reconstruction}
\resizebox{\linewidth}{!}{
\begin{tabular}{@{}lcccccccccc@{}}
\toprule
\multirow{2}{*}{Pipelines} 
& \multicolumn{5}{c}{w/o Decoder Adaptation} 
& \multicolumn{5}{c}{w/ Decoder Adaptation} \\
\cmidrule(lr){2-6} \cmidrule(lr){7-11}
& PSNR ({\color{green!60!black}\bfseries$\pmb{\uparrow}$}) 
& SSIM ({\color{green!60!black}\bfseries$\pmb{\uparrow}$}) 
& LPIPS ({\color{green!60!black}\bfseries$\pmb{\downarrow}$}) 
& IS ({\color{green!60!black}\bfseries$\pmb{\uparrow}$}) 
& FID ({\color{green!60!black}\bfseries$\pmb{\downarrow}$}) 
& PSNR ({\color{green!60!black}\bfseries$\pmb{\uparrow}$}) 
& SSIM ({\color{green!60!black}\bfseries$\pmb{\uparrow}$}) 
& LPIPS ({\color{green!60!black}\bfseries$\pmb{\downarrow}$}) 
& IS ({\color{green!60!black}\bfseries$\pmb{\uparrow}$}) 
& FID ({\color{green!60!black}\bfseries$\pmb{\downarrow}$}) \\
\midrule
SR ($\eta_t = 0.0$)  & 27.64 & 73.1 & 0.067 & 208.53 & 0.73 & 26.72 & 70.2 & 0.077 & 200.00 & 0.85 \\
GAR ($\eta_t = 0.1$) & 26.77 & 70.8 & 0.080 & 204.92 & 0.90 & 26.17 & 68.7 & 0.084 & 197.29 & 0.93 \\
GAR ($\eta_t = 0.2$) & 25.23 & 66.3 & 0.108 & 198.00 & 1.28 & 24.96 & 65.2 & 0.107 & 193.30 & 1.10 \\
GAR ($\eta_t = 0.3$) & 23.72 & 61.1 & 0.146 & 187.94 & 1.88 & 23.59 & 60.6 & 0.141 & 185.83 & 1.44 \\
GAR ($\eta_t = 0.4$) & 22.28 & 55.6 & 0.195 & 172.56 & 2.90 & 22.18 & 55.2 & 0.187 & 173.58 & 2.04 \\
GAR ($\eta_t = 0.5$) & 20.83 & 49.9 & 0.259 & 154.66 & 4.38 & 20.73 & 49.3 & 0.251 & 155.10 & 3.10 \\
\bottomrule
\end{tabular}
}
\vspace{-1ex}
\end{table}

\begin{table*}[!t]
\centering
\caption{\small{\textbf{System-level comparison on class-conditional ImageNet 256${\times}$256}. \textbf{NFE} denotes the number of function evaluations during inference. All numbers are with CFG when applicable, and $\times 2$ in NFE indicates that the CFG computation doubles NFEs at inference time. $^{\dagger}$: results cited from iMF~\citep{Geng2025ImprovedMF}.
}}
\vspace{-2ex}
\resizebox{0.32\linewidth}{!}{
\setlength{\tabcolsep}{4pt}
\small
\begin{tabular}{y{88}x{34}x{24}x{24}}
\toprule
{Method} & {\# Params} & NFE & {FID} \\
\midrule
\rowcolor[gray]{0.9}
\multicolumn{4}{l}{\textit{\textbf{1-NFE diffusion/flow from scratch}}} \\
MeanFlow-XL/2$^{\dagger}$ & 676M & 1 & 3.43   \\
TiM-XL/2$^{\dagger}$ & 664M & 1 & 3.26 \\
$\alpha$-Flow-XL/2+$^{\dagger}$ & 676M & 1 & 2.58 \\
\midrule
MeanFlow-XL/2 + DA & 676M & 1 &  2.96  \\
\midrule
iMF-B/2$^{\dagger}$ & 89M  & 1 & 3.39             \\
iMF-M/2$^{\dagger}$ & 174M & 1 & 2.27             \\
iMF-L/2$^{\dagger}$ & 409M & 1 & 1.86             \\
iMF-XL/2$^{\dagger}$ & 610M & 1 & 1.72    \\
\midrule
iMF-B/2 + DA (Our)    & 89M  & 1 & 2.81          \\
iMF-M/2+ DA (Our)    & 174M & 1 &  1.98         \\
iMF-L/2+ DA (Our)     & 409M & 1 & 1.58           \\
iMF-XL/2+DA (Our)   & 610M & 1 &   \textbf{1.50}  \\
\midrule
\rowcolor[gray]{0.9}
\multicolumn{4}{l}{\textit{\textbf{2-NFE diffusion/flow from scratch}}} \\
IMM-XL/2$^{\dagger}$ & 675M & 1$\times$2 & 7.77 \\
MeanFlow-XL/2+$^{\dagger}$ & 676M & 2 & 2.20  \\
$\alpha$-Flow-XL/2+$^{\dagger}$ & 676M & 2 & 1.95\\
\bottomrule
\end{tabular}
}
\hfill
\resizebox{0.32\linewidth}{!}{
\setlength{\tabcolsep}{4pt}
\small
\begin{tabular}{y{92}x{34}x{30}x{18}}
\toprule
{Method} & {\# Params} & NFE & FID \\
\midrule
\rowcolor[gray]{0.9}
\multicolumn{4}{l}{\textit{\textbf{2-NFE diffusion/flow from scratch}}} \\
iMF-XL/2$^{\dagger}$ & 610M & 2 & 1.54   \\
\midrule
\multicolumn{4}{l} {\textit{\textbf{1-NFE diffusion/flow (distillation)}}} \\
$\pi$-Flow-XL/2$^{\dagger}$ & 675M & 1 & 2.85\\
{DMF-XL/2+}$^{\dagger}$ & 675M &  1 & 2.16\\
FACM-XL/2$^{\dagger}$ & 675M & 1 & 1.76 \\
\midrule
{DMF-XL/2+} + DA (Our) & 676M & 1 &  1.90  \\
\midrule
\multicolumn{4}{l} {\textit{\textbf{Multi-NFE diffusion/flow}}} \\
ADM-G$^{\dagger}$ & 554M & 250$\times$2 & 4.59 \\
LDM-4-G$^{\dagger}$ & 400M & 250$\times$2 & 3.60 \\
SimDiff$^{\dagger}$  & 2B & 1000$\times$2 & 2.77 \\
DiT-XL/2$^{\dagger}$ & 675M & 250$\times$2 & 2.27\\
SiT-XL/2$^{\dagger}$ & 675M   & 250$\times$2 & 2.06  \\
SiT-XL/2\;+\;REPA$^{\dagger}$ & 675M & 250$\times$2 & 1.42 \\
SiD2$^{\dagger}$ & -- & 512$\times$2 & 1.38 \\
LightningDiT-XL/2$^{\dagger}$ & 675M & 250$\times$2 & 1.35 \\
DDT-XL/2$^{\dagger}$ & 675M  & 250$\times$2 & 1.26 \\
\bottomrule
\end{tabular}
}
\hfill
\resizebox{0.32\linewidth}{!}{
\setlength{\tabcolsep}{4pt}
\textcolor[rgb]{0.6,0.6,0.6}{
\small
\begin{tabular}{y{80}x{34}x{30}x{20}}
\toprule
{Method} & {\# Params} & {NFE} & FID \\
\midrule
\multicolumn{4}{l} {\textit{\textbf{Multi-NFE diffusion/flow}}} \\
RAE\;+\;DiT$^\text{DH}$-XL$^{\dagger}$ & 839M & 250$\times$2 & 1.13 \\
\midrule
\multicolumn{4}{l}{\textit{\textbf{GANs}}} \\
BigGAN$^{\dagger}$  & 112M & 1 & 6.95 \\
GigaGAN$^{\dagger}$ & 569M & 1 & 3.45 \\
StyleGAN-XL$^{\dagger}$ & 166M & 1 & 2.30 \\
\midrule
\multicolumn{4}{l}{\textit{\textbf{autoregressive/masking}}} \\
JetFormer-L$^{\dagger}$ & 2.75B & 256$\times$2 & 6.64 \\
MaskGIT$^{\dagger}$ & 227M & 8 & 6.18 \\
RQ-Transformer$^{\dagger}$ & 3.8B & 256$\times$2 & 3.80 \\
STARFlow$^{\dagger}$  & 1.4B & 1024$\times$2 & 2.40 \\
LLamaGen-3B$^{\dagger}$ & 3.1B & 256$\times$2 & 2.18 \\
VAR-$d30$$^{\dagger}$  & 2B   & 10$\times$2 & 1.92 \\
MAR-H$^{\dagger}$  & 943M & 256$\times$2 & 1.55 \\
RAR-XXL$^{\dagger}$ & 1.5B & 256$\times$2 & 1.48\\
xAR-H$^{\dagger}$ & 1.1B & 50$\times$2 & 1.24 \\
\bottomrule
\end{tabular}%
}
}
\vspace{-.5em}
\label{tab:all_results}
\vspace{-4mm}
\end{table*}

\begin{figure}[!h]
    \centering
    \includegraphics[width=\linewidth]{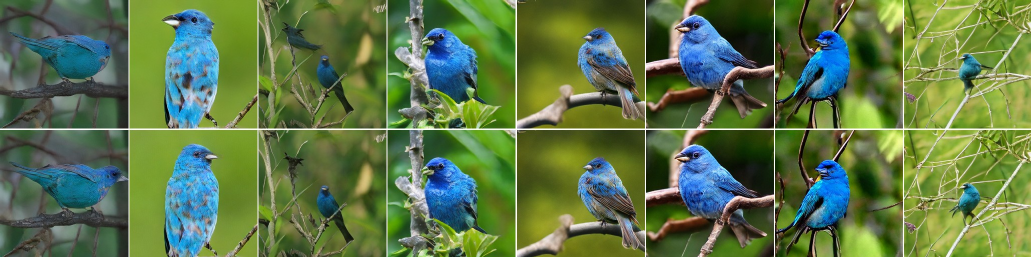}
    \vspace{-4.5ex}
    \caption{Class 14: indigo bunting (top: before, bottom: after).}
    \label{fig:appendix_da_class_014}
    \vspace{-1ex}
\end{figure}

\begin{figure}[!h]
    \centering
    \includegraphics[width=\linewidth]{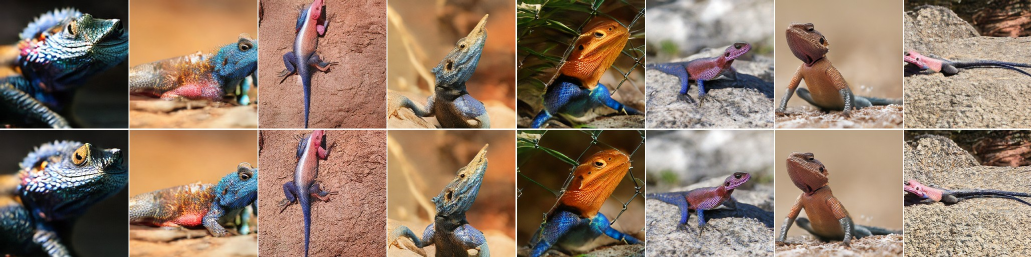}
    \vspace{-4.5ex}
    \caption{Class 42: agama (top: before, bottom: after).}
    \label{fig:appendix_da_class_042}
    \vspace{-1ex}
\end{figure}

\begin{figure}[!h]
    \centering
    \includegraphics[width=\linewidth]{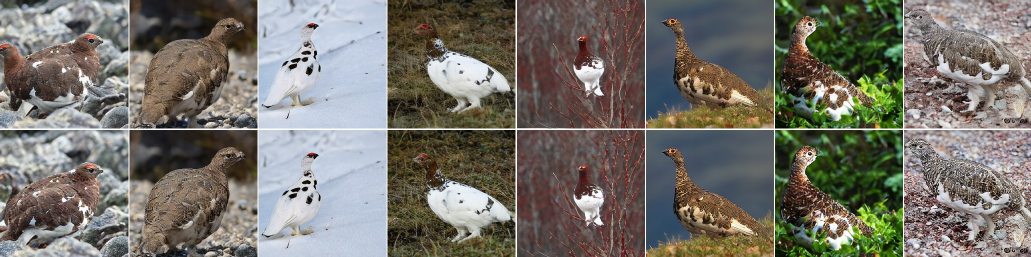}
    \vspace{-3ex}
    \caption{Class 81: ptarmigan (top: before, bottom: after).}
    \label{fig:appendix_da_class_081}
    \vspace{-1ex}
\end{figure}

\begin{figure}[!h]
    \centering
    \includegraphics[width=\linewidth]{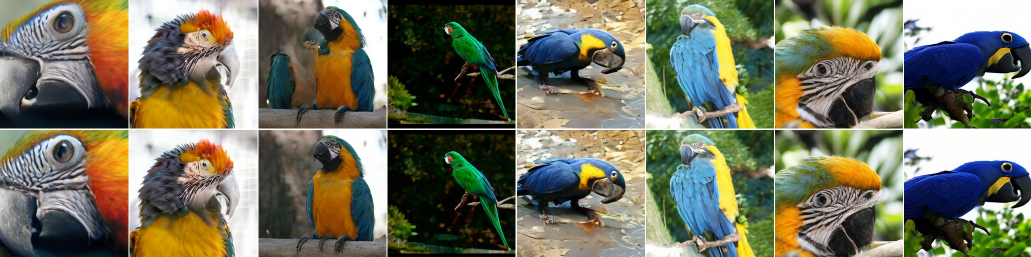}
    \vspace{-4.5ex}
    \caption{Class 88: macaw (top: before, bottom: after).}
    \label{fig:appendix_da_class_088}
    \vspace{-1ex}
\end{figure}

\begin{figure}[!h]
    \centering
    \includegraphics[width=\linewidth]{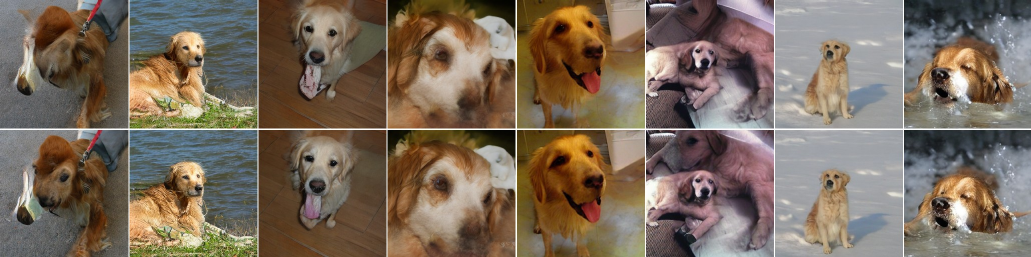}
    \vspace{-4.5ex}
    \caption{Class 207: golden retriever (top: before, bottom: after).}
    \label{fig:appendix_da_class_207}
    \vspace{-1ex}
\end{figure}

\begin{figure}[!h]
    \centering
    \includegraphics[width=\linewidth]{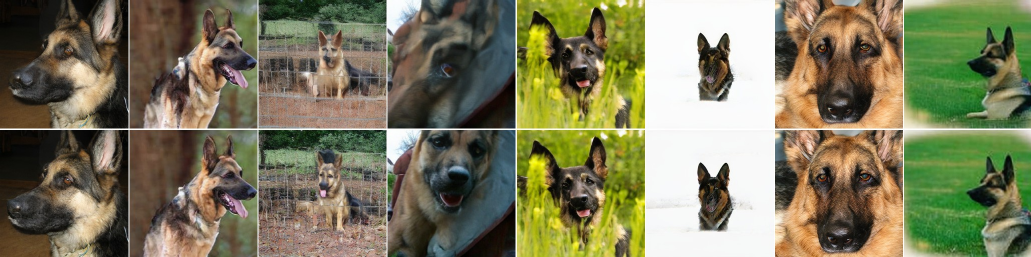}
    \vspace{-4.5ex}
    \caption{Class 235: German shepherd (top: before, bottom: after).}
    \label{fig:appendix_da_class_235}
    \vspace{-1ex}
\end{figure}

\begin{figure}[!h]
    \centering
    \includegraphics[width=\linewidth]{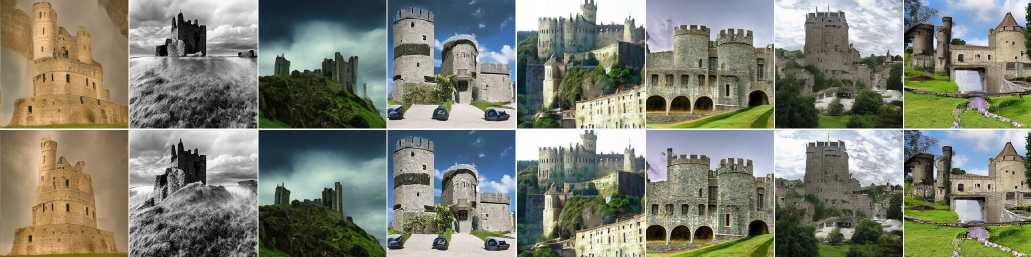}
    \vspace{-4.5ex}
    \caption{Class 483: castle (top: before, bottom: after).}
    \label{fig:appendix_da_class_483}
    \vspace{-1ex}
\end{figure}

\begin{figure}[!h]
    \centering
    \includegraphics[width=\linewidth]{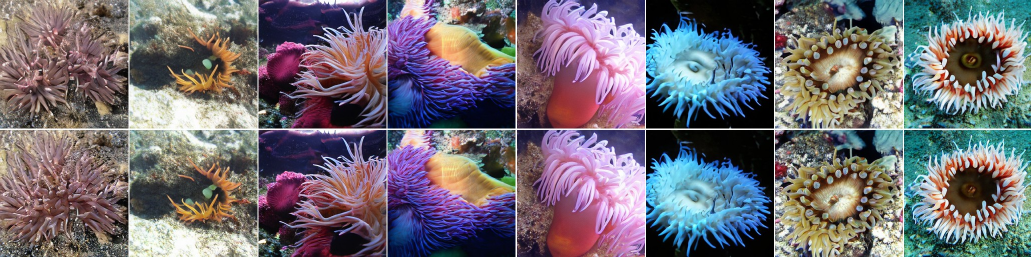}
    \vspace{-4.5ex}
    \caption{Class 108: sea anemone (top: before, bottom: after).}
    \label{fig:appendix_da_class_108}
    \vspace{-1ex}
\end{figure}

\begin{figure}[!h]
    \centering
    \includegraphics[width=\linewidth]{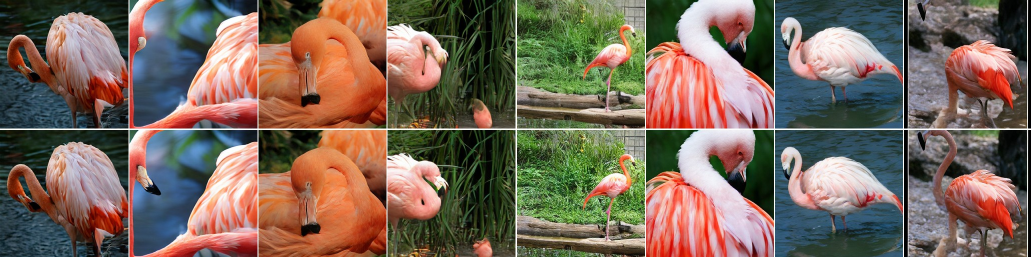}
    \vspace{-4.5ex}
    \caption{Class 130: flamingo (top: before, bottom: after).}
    \label{fig:appendix_da_class_130}
    \vspace{-1ex}
\end{figure}

\begin{figure}[!h]
    \centering
    \includegraphics[width=\linewidth]{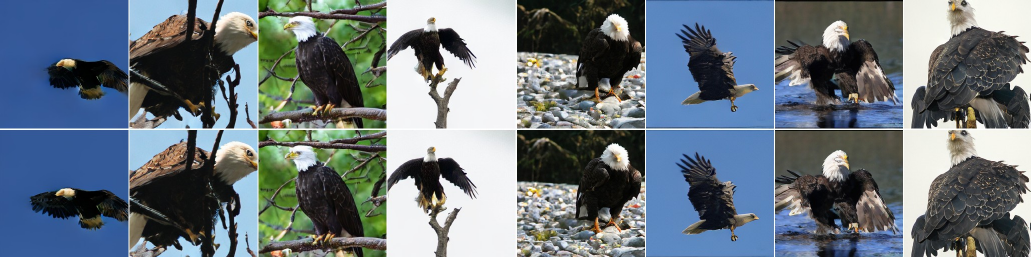}
    \vspace{-4.5ex}
    \caption{Class 22: bald eagle (top: before, bottom: after).}
    \label{fig:appendix_da_class_22}
    \vspace{-1ex}
\end{figure}

\begin{figure}[!h]
    \centering
    \includegraphics[width=\linewidth]{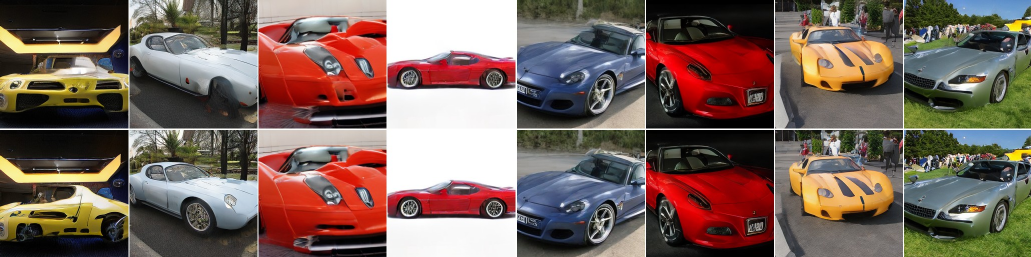}
    \vspace{-4.5ex}
    \caption{Class 817: sports car (top: before, bottom: after).}
    \label{fig:appendix_da_class_817}
    \vspace{-1ex}
\end{figure}

\paragraph{Visual Comparison of Generated Samples.}
We provide additional qualitative comparisons for decoder adaptation (before and after) on iMF-XL/2 across 11 different classes, as shown in Figures~\ref{fig:appendix_da_class_014}-\ref{fig:appendix_da_class_817}. For each class, we present two sets of eight samples: the top row shows results before decoder adaptation, while the bottom row shows results after adaptation. Samples in the same column are generated from identical generative latents $z_g$, with the only difference being the decoder parameters. Across all classes, decoder adaptation consistently produces samples with richer fine-grained details, whereas samples before adaptation tend to appear slightly smoother. In addition, decoder adaptation reduces visual artifacts and yields more semantically consistent samples.

\end{appendices}


\end{document}